%% file: main.tex
\pdfoutput=1
\documentclass[11pt]{article}

\usepackage{acl}

\usepackage{times}
\usepackage{latexsym}

\usepackage[T1]{fontenc}
\usepackage[utf8]{inputenc}

\usepackage{microtype}
\usepackage{balance}

\usepackage{amsmath,amssymb,bm,float,graphicx,multirow,natbib,soul,xspace}
\usepackage{bbm} % indicator function \mathbbm{1}_{[k \ne i]}
\usepackage{booktabs} % tables, toprule, bottomrule
\usepackage{hyperref} % href
\usepackage[mathscr]{eucal}
\usepackage{subcaption} % subfigures
\usepackage{wrapfig} % wrap text around figures and tables

\newcommand{\dmatrix}[1]{\bm{\uppercase{#1}}}
\newcommand{\dvector}[1]{\bm{#1}}

\newcommand{\tldc}{TrLDC\xspace}
\title{Revisiting Transformer-based Models for Long Document Classification}

\author{Xiang Dai~\thanks{\,\,\,This work was partially done when Dai was at the University of Copenhagen.} \\
  CSIRO Data61 \\
  \texttt{dai.dai@csiro.au} \And
  Ilias Chalkidis \\
  University of Copenhagen \\
  \texttt{ilias.chalkidis@di.ku.dk} \AND
  Sune Darkner \\
  University of Copenhagen \\
  \texttt{darkner@di.ku.dk} \And
  Desmond Elliott \\
  University of Copenhagen \\ Pioneer Centre for AI  \\
  \texttt{de@di.ku.dk} \\}

\begin{document}
\maketitle

%\input{response}
\input{body}

% Entries for the entire Anthology, followed by custom entries
\bibliography{main}
\bibliographystyle{acl_natbib}

\clearpage

\input{appendix}

\end{document}

%% file: body.tex
%%%%%%%%%%%%%%%%%%%%%%%%%%%%%%%%%%%%   0411
\begin{abstract}
The recent literature in text classification is biased towards short text sequences (e.g., sentences or paragraphs). 
In real-world applications, multi-page multi-paragraph documents are common and they cannot be efficiently encoded by vanilla Transformer-based models.  
We compare different Transformer-based Long Document Classification (\tldc) approaches that aim to mitigate the computational overhead of vanilla transformers to encode much longer text, namely sparse attention and hierarchical encoding methods.
We examine several aspects of sparse attention (e.g., size of local attention window, use of global attention) and hierarchical (e.g., document splitting strategy) transformers on four document classification datasets covering different domains. 
We observe a clear benefit from being able to process longer text, and, based on our results, we derive practical advice of applying Transformer-based models on long document classification tasks.
% We find that, there is a clear benefit from being able to process longer text, and if applied properly, Transformer-based models can outperform task-specific CNN architectures for long document classification.
\end{abstract}

%%%%%%%%%%%%%%%%%%%%%%%%%%%%%%%%%%%%   0411
\section{Introduction}
Natural language processing has been revolutionised by the large scale self-supervised pre-training of language encoders ~\cite{devlin-google-2019-naacl-bert,liu-fb-2019-roberta}, which are fine-tuned in order to solve a wide variety of downstream classification tasks.
However, the recent literature in text classification mostly focuses on short sequences, such as sentences or paragraphs~\cite{sun-fudan-2019-ccl-finetune,adhikari-waterloo-2019-docbert,mosbach-saarland-2021-iclr-finetune}, which are sometimes misleadingly named as documents.\footnote{For example, many biomedical datasets use `documents' from the \href{https://pubmed.ncbi.nlm.nih.gov/}{PubMed} collection of biomedical literature, but these documents actually consist of titles and abstracts.}

\begin{figure}[t]
    \centering
    \includegraphics[width=\linewidth]{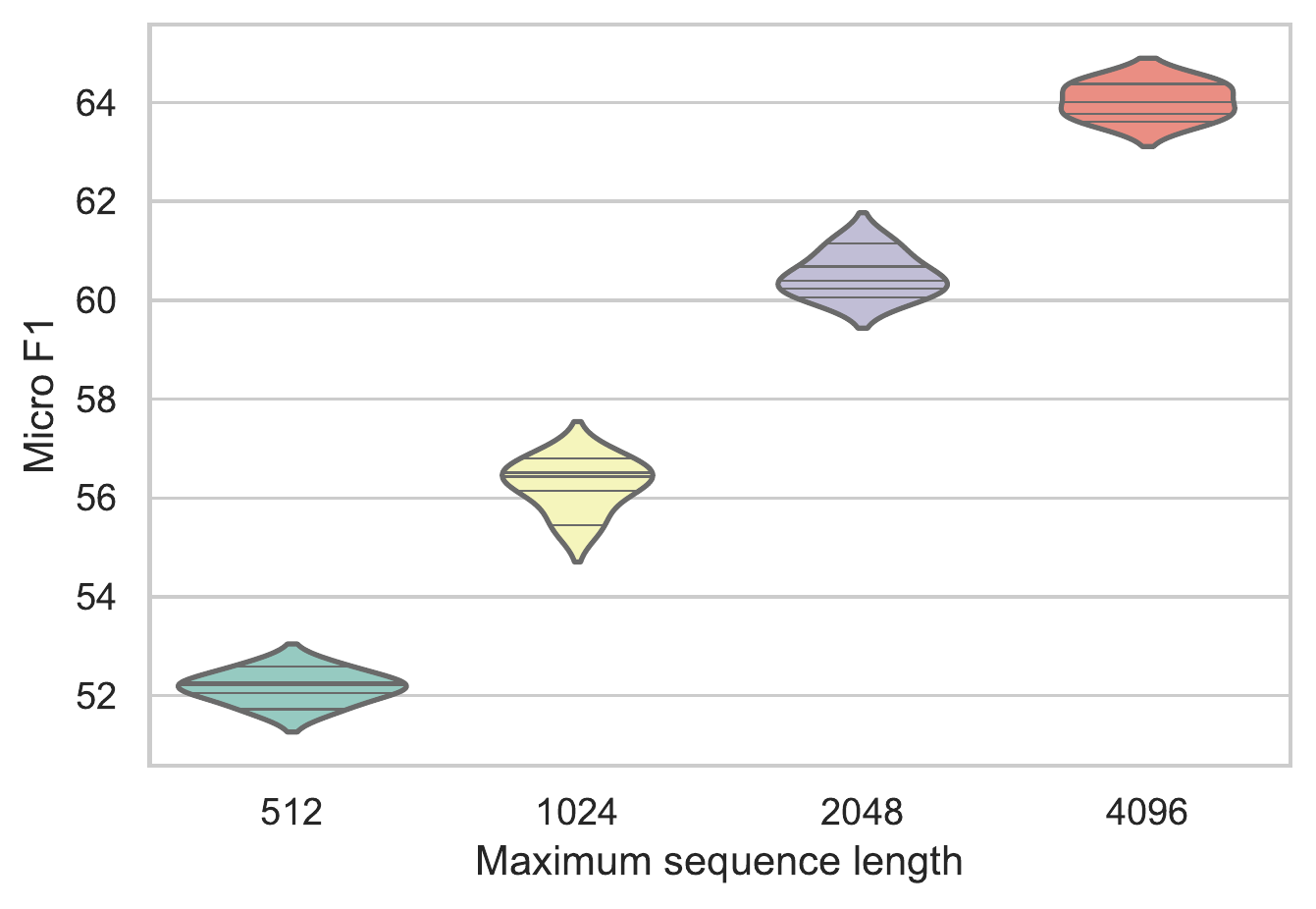}
    \caption{The effectiveness of Longformer, a long-document Transformer, on the MIMIC-III development set. There is a clear benefit from being able to process longer text.}
    \label{figure1_max_sequence_length_vs_f1}
    \vspace{-4mm}
\end{figure}

The transition from short-to-long document classification is non-trivial.
One challenge is that BERT and most of its variants are pre-trained on sequences containing up-to 512 tokens, which is not a long document. 
A common practice is to truncate actually long documents to the first 512 tokens, which allows the immediate application of these pre-trained models~\cite{adhikari-waterloo-2019-docbert,chalkidis-aueb-2020-emnlp-multilabel}.
We believe that this is an insufficient approach for long document classification because truncating the text may omit important information, leading to poor classification performance (Figure~\ref{figure1_max_sequence_length_vs_f1}). 
% See Figure~\ref{figure1_max_sequence_length_vs_f1} for empirical evidence to support this claim.
Another challenge comes from the computational overhead of vanilla Transformer: in the multi-head self-attention operation~\cite{vaswani-google-2017-neurips-transformer}, each token in a sequence of $n$ tokens attends to all other tokens.
This results in a function that has $O(n^2)$ time and memory complexity, which makes it challenging to efficiently process long documents.

In response to the second challenge, long-document Transformers have emerged to deal with long sequences~\cite{beltagy-allenai-2020-longformer,zaheer-google-2020-neurips-bigbird}. 
However, they experiment and report results on non-ideal long document classification datasets, i.e., documents on the IMDB dataset are not really long -- fewer than $15$\% of examples are longer than $512$ tokens; while the Hyperpartisan dataset only has very few ($645$ in total) documents. 
On datasets with longer documents, such as the MIMIC-III dataset~\cite{johnson-mit-2016-mimic-iii} with an average length of 2,000 words, it has been shown that multiple variants of BERT perform worse than a CNN or RNN-based model~\cite{chalkidis-aueb-2020-emnlp-multilabel,vu-csiro-2020-ijcai-laat,dong-ed-2021-jbi-icd-coding,ji-aalto-2021-icd-coding,gao-ornl-2021-jbhi-icd-coding,pascual-ethz-2021-bionlp-icd-coding}. 
We believe there is a need to understand the performance of Transformer-based models on classifying documents that are actually long.

In this work, we aim to transfer the success of the pre-train--fine-tune paradigm to long document classification.
Our main contributions are:
\begin{itemize}
    \item We compare different long document classification approaches based on transformer architecture: namely, sparse attention, and hierarchical methods. 
    Our results show that processing more tokens can bring drastic improvements comparing to processing up-to 512 tokens.
    \item We conduct careful analyses to understand the impact of several design options on both the effectiveness and efficiency of different approaches.
    Our results show that some design choices (e.g., size of local attention window in sparse attention method) can be adjusted to improve the efficiency without sacrificing the effectiveness, whereas some choices (e.g., document splitting strategy in hierarchical method) vastly affect effectiveness.
    \item Last but not least, our results show that, contrary to previous claims, Transformer-based models can outperform former state-of-the-art CNN based models on MIMIC-III dataset .
    %\item Based on our empirical results, we derive practical advice of applying Transformer-based models to long document classification.
    %on two challenging datasets from clinical and legal domains show that several design approaches (e.g., the length of local attention windows) vastly affect results.
    %\item We also investigate the impact of over-segmentation, an issue to which was attributed by recently published negative results of applying transformer-based models on domain-specific datasets. Our results show that XXX simply keeping the first 2 wordpieces of these over-segmented words cause neglectable performance drop XXX.
\end{itemize}

%%%%%%%%%%%%%%%%%%%%%%%%%%%%%%%%%%%%   0411
\section{Problem Formulation and Datasets}

We divide the document classification model into two components: (1) a document encoder, which builds a vector representation of a given document; and, (2) a classifier that predicts a single or multiple labels given the encoded vector.
In this work, we mainly focus on the first component: we use Transformer-based encoders to build a document representation, and then take the encoded document representation as the input to a classifier.
For the second component, we use a \textsc{Tanh} activated hidden layer, followed by the output layer.
Output probabilities are obtained by applying a \textsc{Sigmoid} (multi-label) or \textsc{Softmax} (multi-class) function to output logits.\footnote{Long document classification datasets are usually annotated using a large number of labels. Studies that have focused on the second component investigate methods of utilising label hierarchy~\cite{chalkidis-aueb-2020-emnlp-multilabel,vu-csiro-2020-ijcai-laat}, pre-training label embeddings~\cite{dong-ed-2021-jbi-icd-coding}, to name but a few.}

\begin{figure}[t]
    \centering
    \includegraphics[width=\linewidth]{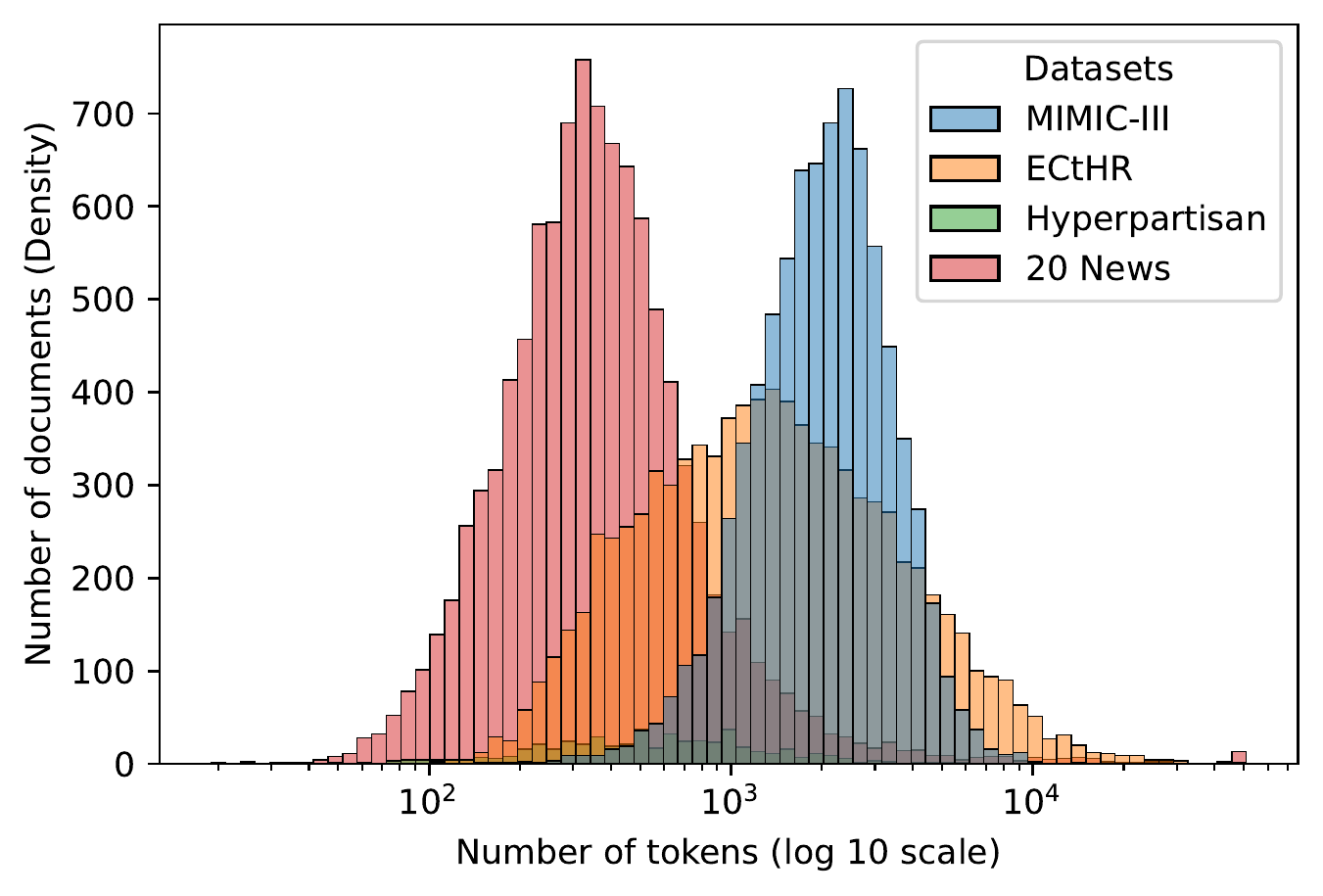}
    \caption{The distribution of document lengths. A log-$10$ scale is used for the X axis.}
    \label{figure-sequence-length-distribution}
    \vspace{-4mm}
\end{figure}

We mainly conduct our experiments on the MIMIC-III dataset~\cite{johnson-mit-2016-mimic-iii}, where researchers still fail to transfer ``the Magic of BERT'' to medical code assignment tasks~\cite{ji-aalto-2021-icd-coding,pascual-ethz-2021-bionlp-icd-coding}.
\begin{description}
\item[MIMIC-III] contains Intensive Care Unit (ICU) discharge summaries, each of which is annotated with multiple labels---\emph{diagnoses} and \emph{procedures}---using the ICD-9 (The International Classification of Diseases, Ninth Revision) hierarchy.
Following~\citet{mullenbach-gatech-2018-naacl-caml}, we conduct experiments using the top 50 frequent labels.\footnote{Details about dataset split and labels can be found at \href{https://github.com/jamesmullenbach/caml-mimic/blob/master/notebooks/dataproc_mimic_III.ipynb}{https://github.com/jamesmullenbach/caml-mimic}}
\end{description}

To address the generalisation concern, we also use three datasets from other domains: ECtHR~\cite{chalkidis-jana-2022-acl-lexglue} sourced from legal cases, Hyperpartisan~\cite{kiesel-mestre-2019-semeval-hyperpartisan} and 20 News~\cite{joachims-1997-icml-20news}, both from news articles.
\begin{description}
\item[ECtHR] contains legal cases from The European Court of Human Rights' public database. The court hears allegations that a state has breached human rights provisions of the European Convention of Human Rights, and each case is mapped to one or more \emph{articles} of the convention that were \emph{allegedly} violated.\footnote{\href{https://huggingface.co/datasets/ecthr_cases}{https://huggingface.co/datasets/ecthr\_cases}}
\item[Hyperpartisan] contains news articles which are manually labelled as hyperpartisan (taking an extreme left or right standpoint) or not.\footnote{\href{https://pan.webis.de/semeval19/semeval19-web/}{https://pan.webis.de/semeval19/semeval19-web/}; we use the \href{https://github.com/allenai/longformer/blob/master/scripts/hp-splits.json}{split} provided by~\citet{beltagy-allenai-2020-longformer}.}
\item[20 News] contains newsgroups posts which are categorised into 20 topics.\footnote{\href{http://qwone.com/~jason/20Newsgroups/}{http://qwone.com/\string~jason/20Newsgroups/}}
\end{description}
We note that documents in MIMIC-III and ECtHR are much longer than those in Hyperpartisan and 20 News (Table~\ref{table-data-statistics} in Appendix and Figure~\ref{figure-sequence-length-distribution}).

%%%%%%%%%%%%%%%%%%%%%%%%%%%%%%%%%%%%   0410
\section{Approaches}
\label{sec:approaches}
In the era of Transformer-based models, we identify two representative approaches of processing long documents in the literature that either acts as an inexpensive drop-in replacement for the vanilla self-attention (i.e., \emph{sparse} attention) or builds a task-specific architecture (i.e., \emph{hierarchical} Transformers).

\subsection{Sparse-Attention Transformers}

Vanilla transformer relies on the multi-head self-attention mechanism, which scales poorly with the length of the input sequence, requiring quadratic computation time and memory to store all scores that are used to compute the gradients during back-propagation~\cite{qiu-fb-2020-emnlp-blockbert}. 
Several Transformer-based  models~\cite{kitaev-google-2020-iclr-reformer,tay-google-2020-efficient-transformers,choromanski-google-2021-iclr-performers} have been proposed exploring efficient alternatives that can be used to process long sequences.

\begin{figure}[t]
    \centering
    \includegraphics[width=0.95\linewidth]{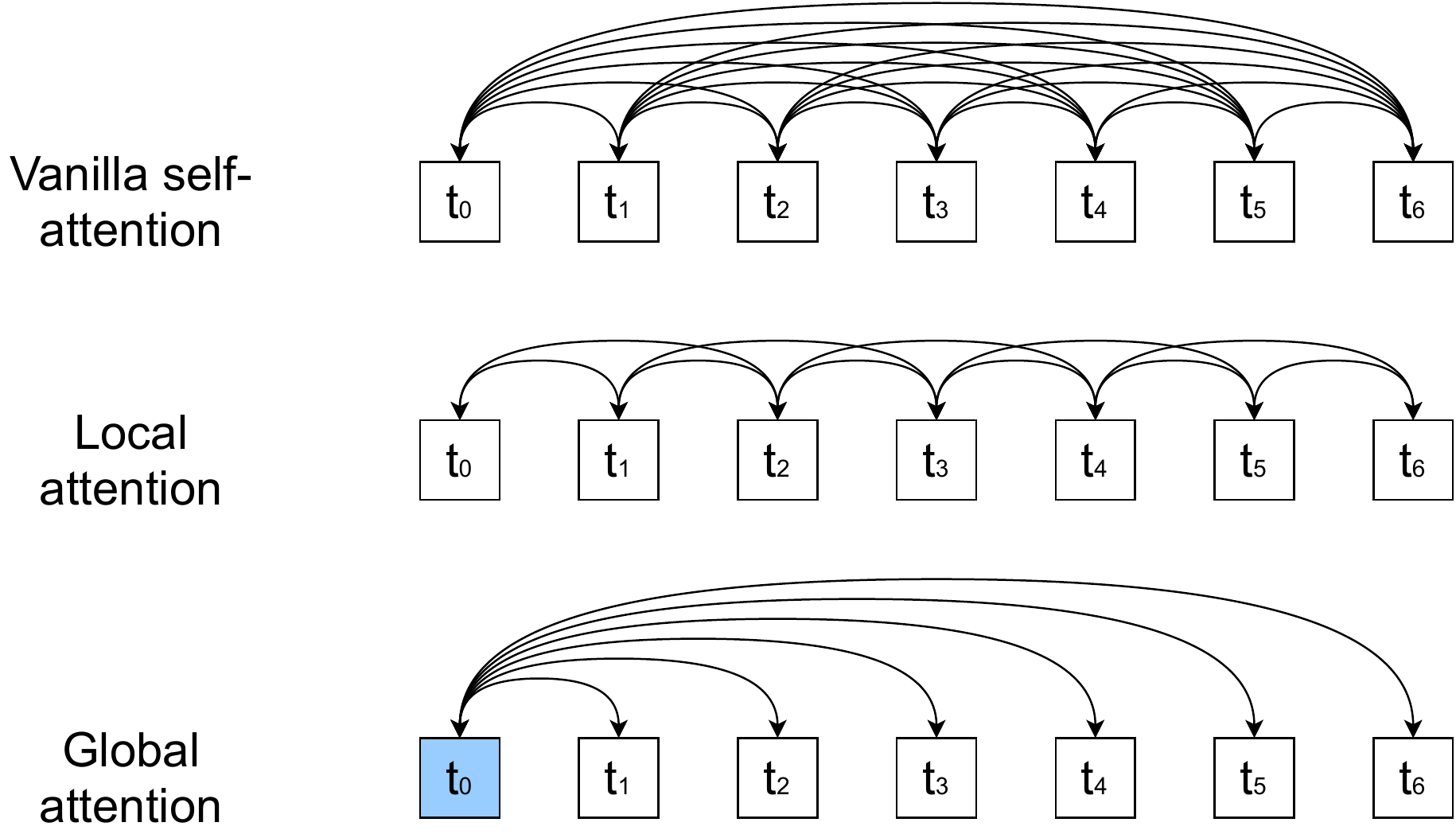}
    \caption{A comparison of three types of attention operations. The example sequence contains 7 tokens; we set local attention window size as 2, and only the first token using global attention. Note that these curves are bi-directional that tokens can attend to each other.}
    \label{figure-longformer-illustration}
    \vspace{-4mm}
\end{figure}

\textbf{Longformer} of \citet{beltagy-allenai-2020-longformer} consists of local (window-based) attention and global attention that reduces the computational complexity of the model and thus can be deployed to process up to $4096$ tokens. Local attention is computed in-between a window of neighbour (consecutive) tokens. Global attention relies on the idea of global tokens that are able to attend and be attended by any other token in the sequence (Figure~\ref{figure-longformer-illustration}).
\textbf{BigBird} of \citet{zaheer-google-2020-neurips-bigbird} is another sparse-attention based Transformer that uses a combination of a local, global and random attention, i.e., all tokens also attend a number of random tokens on top of those in the same neighbourhood. 
Both models are warm-started from the public RoBERTa checkpoint and are further pre-trained on masked language modelling. They have been reported to outperform RoBERTa on a range of tasks that require modelling long sequences.

We choose Longformer~\cite{beltagy-allenai-2020-longformer} in this study and refer readers to~\citet{xiong-fb-2021-local-attention} for a systematic comparison of recent proposed efficient attention variants.

\begin{figure}[t]
    \centering
    \includegraphics[width=0.95\linewidth]{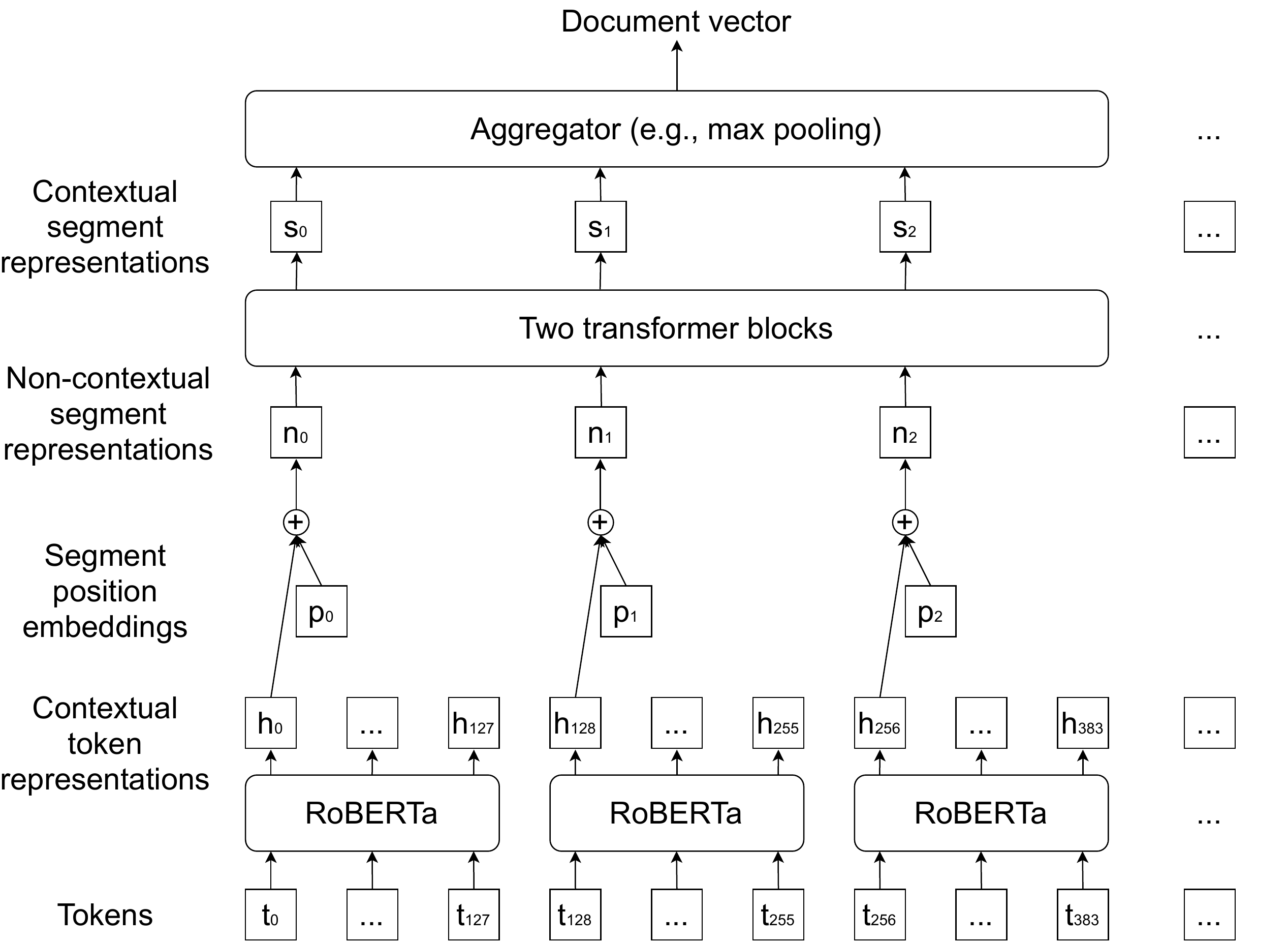}
    \caption{A high-level illustration of hierarchical Transformers. A shared pre-trained RoBERTa is used to encode each segment, and a two layer transformer blocks is used to capture the interaction between different segments. Finally, contextual segment representations are aggregated into a document representation.}
    \label{figure-hierarchical-transformer}
    \vspace{-4mm}
\end{figure}

\subsection{Hierarchical Transformers}

Instead of modifying multi-head self-attention mechanism to efficiently model long sequences, hierarchical Transformers build on top of vanilla transformer architecture.

A document, $\mathcal{D} = \{t_0, t_1, \cdots, t_{|\mathcal{D}|}\}$, is first split into segments, each of which should have less than $512$ tokens.
These segments can be independently encoded using any pre-trained Transformer-based encoders (e.g., RoBERTa in Figure~\ref{figure-hierarchical-transformer}).
We sum the contextual representation of the first token from each segment up with segment position embeddings as the segment representation (i.e., $n_i$ in Figure~\ref{figure-hierarchical-transformer}).
Then the segment encoder---two transformer blocks~\cite{zhang-microsoft-2019-acl-hierarchical}---are used to capture the interaction between segments and output a list of contextual segment representations (i.e., $s_i$ in Figure~\ref{figure-hierarchical-transformer}), which are finally aggregated into a document representation.
By default, the aggregator is the \textbf{max-pooling} operation unless other specified.\footnote{Code is available at URL} %\href{https://github.com/coastalcph/trldc}{https://github.com/coastalcph/trldc}}

\section{Experimental Setup}
\paragraph{Backbone Models}
We mainly consider two models in our experiments: Longformer-base~\cite{beltagy-allenai-2020-longformer}, and RoBERTa-base~\cite{liu-fb-2019-roberta} which is used in hierarchical Transformers.
% similar to SMITH of \cite{yang-zhang-2020-cikm}.

\paragraph{Evaluation metrics}
For the MIMIC-III (multilabel) dataset, we follow previous work~\cite{mullenbach-gatech-2018-naacl-caml,cao-ac-2020-acl-icd-coding} and use micro-averaged AUC (Area Under the receiver operating characteristic Curve), macro-averaged AUC, micro-averaged $F_1$, macro-averaged $F_1$ and Precision@5---the proportion of the ground truth labels in the top-5 predicted labels---as the metrics. 
We report micro and macro averaged $F_1$ for the ECtHR (multilabel) dataset, and accuracy for both Hyperpartisan (binary) and 20 News (multiclass) datasets.

\section{Experiments}
We conduct a series of controlled experiments to understand the impact of design choices in different \tldc models.
Bringing these optimal choices all together, we compare \tldc against the state of the art, as well as baselines that only process up-to 512 tokens.
Finally, based on our investigation, we derive practical advice of applying transformer-based models to long document classification regarding both effectiveness and efficiency.
% We show, contrary to previously-reported results, that the benefits of pre-trained Transformers also apply to long document classification.

\begin{figure}[t]
\begin{subfigure}{.5\linewidth}
  \centering
  \includegraphics[width=\textwidth]{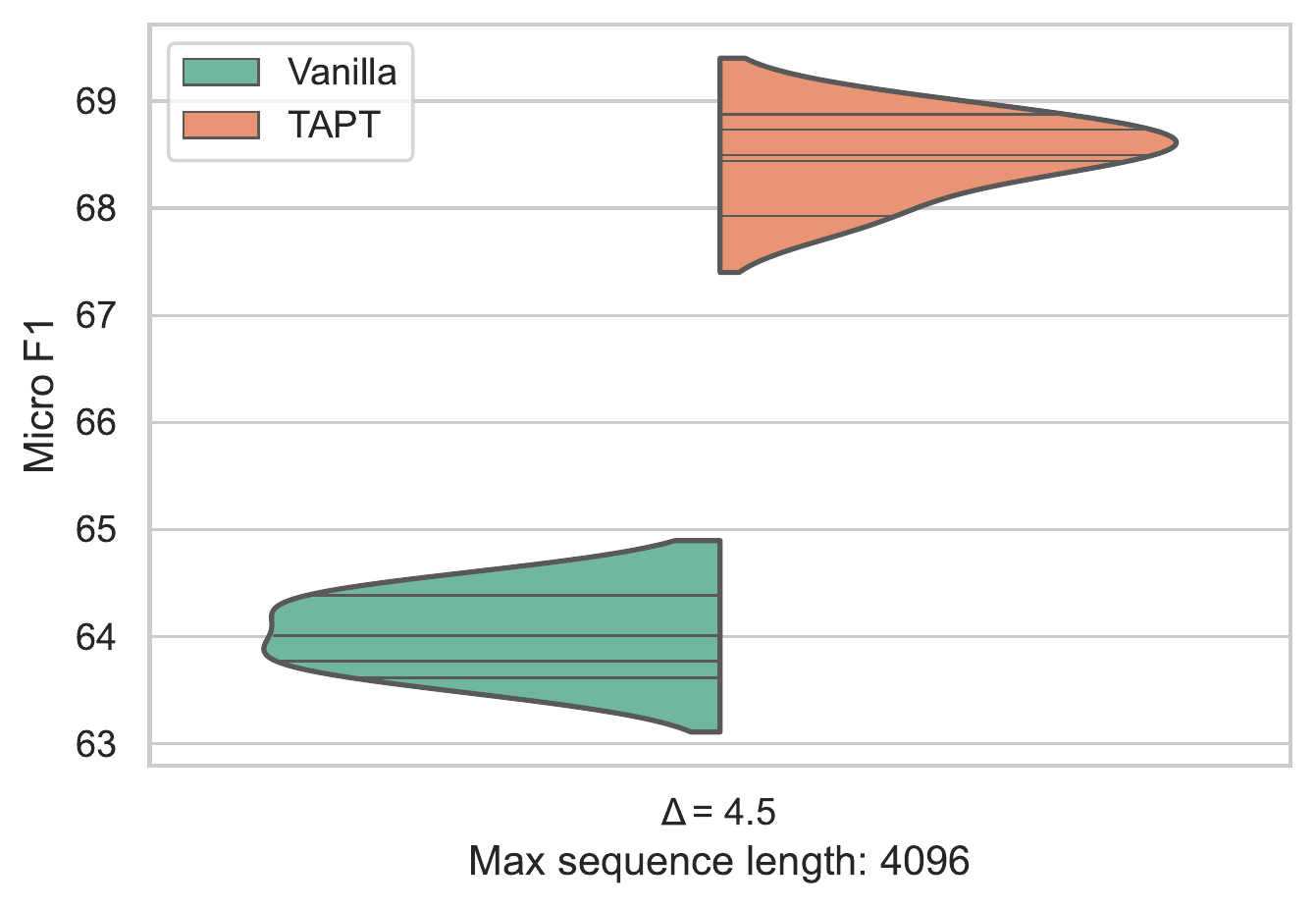}
  \caption{Longformer on MIMIC-III}
\end{subfigure}%
\begin{subfigure}{.5\linewidth}
  \centering
  \includegraphics[width=\textwidth]{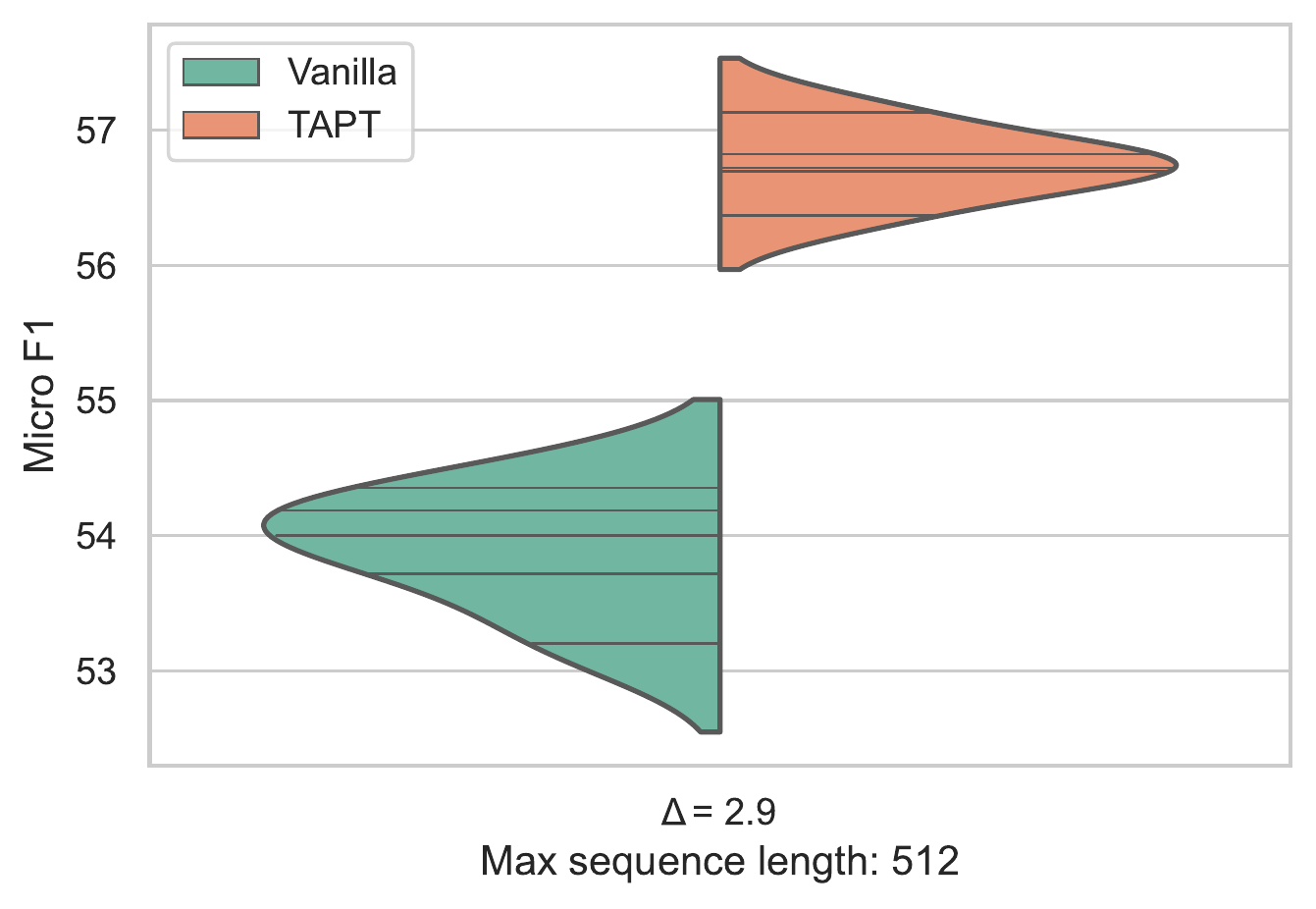}
  \caption{RoBERTa on MIMIC-III}
\end{subfigure}
\\[2ex]
\begin{subfigure}{.5\linewidth}
  \centering
  \includegraphics[width=\textwidth]{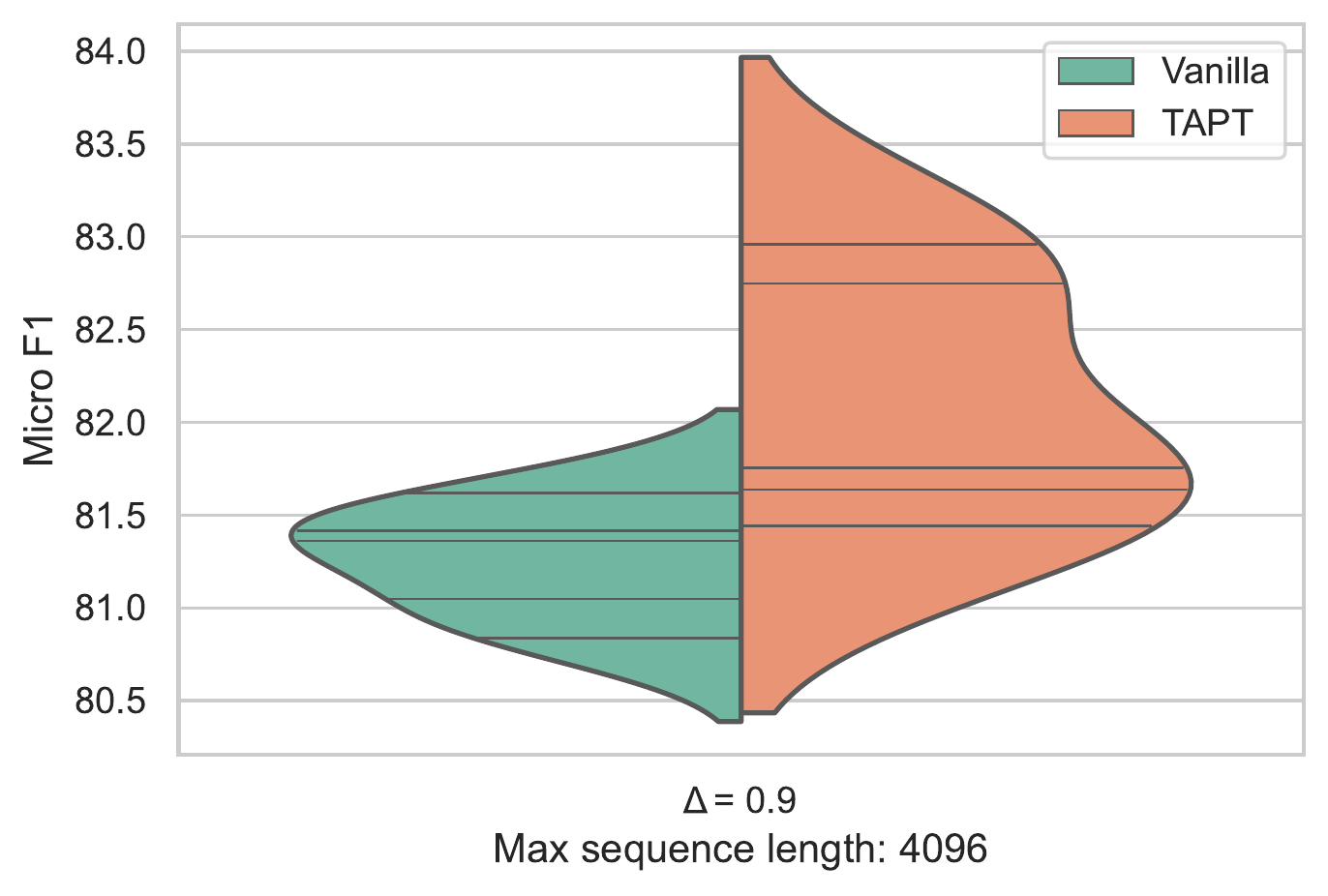}
  \caption{Longformer on ECtHR}
\end{subfigure}%
\begin{subfigure}{.5\linewidth}
  \centering
  \includegraphics[width=\textwidth]{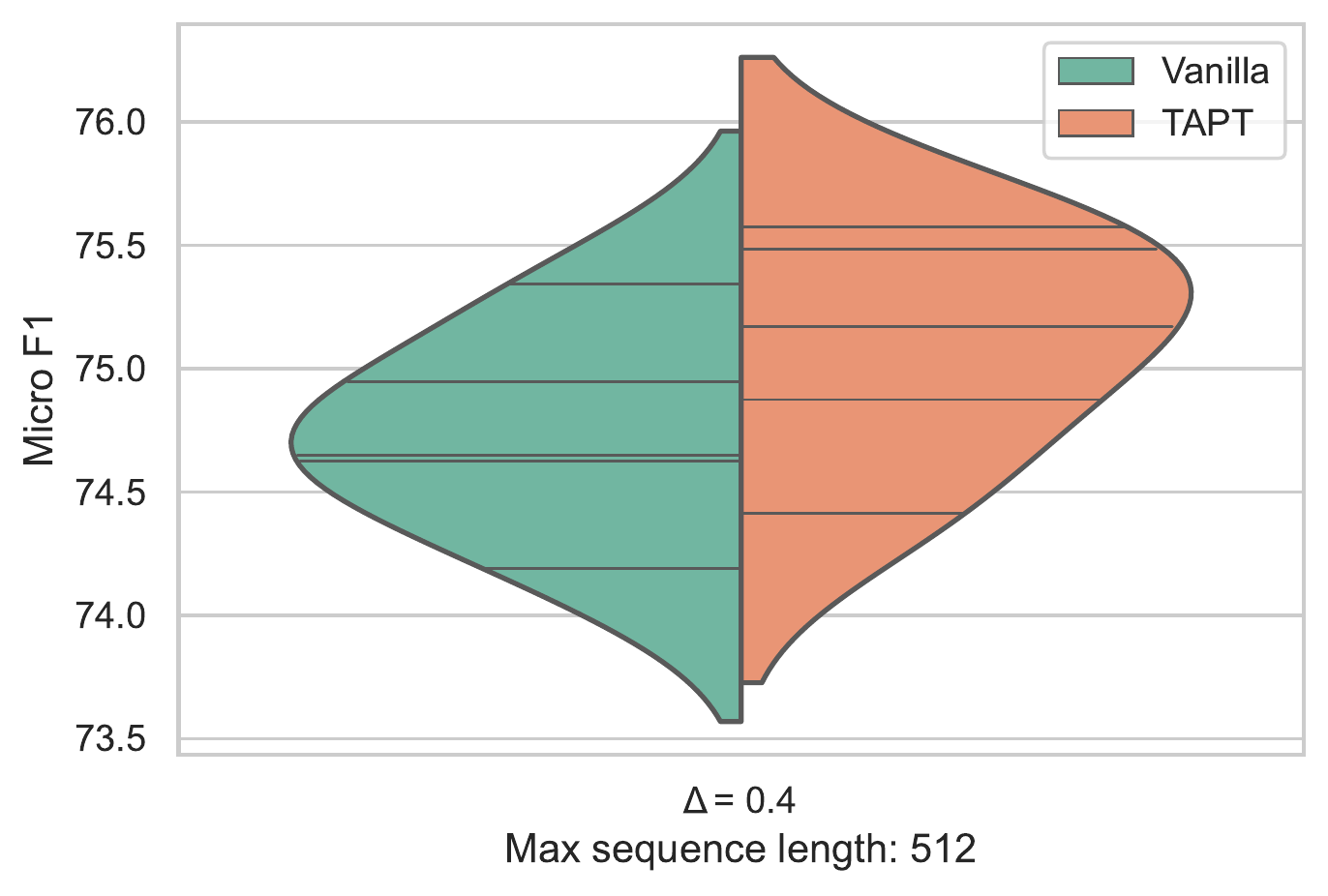}
  \caption{RoBERTa on ECtHR}
\end{subfigure}
\caption{Task-adaptive pre-training (right side in each plot) can improve the effectiveness (measured on the development sets) of pre-trained models by a large margin on MIMIC-III, but small on ECtHR. $\Delta$: the difference between mean values of compared experiments.}
\label{figure-tapt-results}
\vspace{-4mm}
\end{figure}

\paragraph{Task-adaptive pre-training is a promising first step.}\hspace{-2ex}
Domain-adaptive pre-training (DAPT) -- the continued pre-training a language model on a large corpus of domain-specific text -- is known to improve downstream task performance~\cite{gururangan-allenai-2020-acl-dapt,jorgensen-ku-2021-emnlp-mdapt}.
%  in many recent work and has become common practices in the literature when task data are from a specific domain
However, task-adaptive pre-training (TAPT) -- continues unsupervised pre-training on the task's data -- is comparatively less studied, mainly because most of the benchmarking corpora are small and thus the benefit of TAPT seems less obvious than DAPT.
% For example, \citet{gururangan-marasovic-2020-acl} observe a absolute improvement of $2.3$ using DAPT on ChemProt dataset, whereas an improvement of $0.7$ using TAPT.

We believe document classification datasets, due to their relatively large size, can benefit from TAPT.
On both MIMIC-III and ECtHR, we continue to pre-train Longformer and RoBERTa using the masked language modelling pre-training objective (details about pre-training can be found at Appendix~\ref{appendix-section-pretraining-details}).
We find that task-adaptive pre-trained models substantially improve performance on MIMIC-III (Figure~\ref{figure-tapt-results} (a) and (b)), but there are smaller improvements on ECtHR (Figure~\ref{figure-tapt-results} (c) and (d)). 
We suspect this difference is because legal cases (i.e., ECtHR) are publicly available and have been covered in pre-training data used for training Longformer and RoBERTa, whereas clinical notes (i.e., MIMIC-III) are not \cite{dodge-allenai-2021-emnlp-corpora}.
See Appendix~\ref{appendix-section-clinical-note-and-legal-document} for a short analysis on this matter.

We also compare our TAPT-RoBERTa against publicly available domain-specific RoBERTa, trained from scratch on biomedical articles and clinical notes.
Results (Figure~\ref{appendix-figure-compare-roberta-version} in Appendix) show that TAPT-RoBERTa outperforms domain-specific base model, but underperforms the larger model.
\vspace{1mm}

\subsection{Longformer}
\begin{table}[t!]
    \centering
    \begin{tabular}{c c cc}
    \toprule
    \multirow{2}{*}{Size} & \multirow{2}{*}{Micro $F_1$} & \multicolumn{2}{c}{Speed} \\
     & & Train & Test \\
    \midrule
32 & 67.9 \tiny $\pm$ 0.3 & 9.9 (2.9x) & 45.6 (2.8x) \\ % 2022-02-07-A
64 & 68.1 \tiny $\pm$ 0.1 & 8.8 (2.6x) & 41.4 (2.5x) \\ % 2022-02-07-A
128 & 68.3 \tiny $\pm$ 0.3 & 7.4 (2.1x) & 34.1 (2.1x) \\ % 2022-02-07-A
256 & 68.4 \tiny $\pm$ 0.3 & 5.5 (1.6x) & 25.4 (1.6x) \\ % 2022-02-07-A
512 & 68.5 \tiny $\pm$ 0.3 & 3.5 (1.0x) & 16.3 (1.0x) \\ % 2022-02-07-A
    \bottomrule
    \end{tabular}
    \caption{The impact of local attention window size in Longformer on MIMIC-III development set. Speed is measured using `processed samples per second', and numbers in parenthesis are the relative speedup.}
    \label{table-window-in-Longformer}
    \vspace{-4mm}
\end{table}

\paragraph{Small local attention windows are effective and efficient.}\hspace{-2ex} \citet{beltagy-allenai-2020-longformer} observe that many tasks do not require reasoning over the entire context.
For example, they find that the distance between any two mentions in a coreference resolution dataset (i.e., OntoNotes) is small, and it is possible to achieve competitive performance by processing small segments containing these mentions.
%segmenting a long document into smaller segments and then processing them without considering cross segment interactions.

Inspired by this observation, we investigate the impact of local context size on document classification, regarding both effectiveness and efficiency.
We hypothesise that long document classification, which is usually paired with a large label space, can be performed by models that only attend over short sequences instead of the entire document~\cite{gao-ornl-2021-jbhi-icd-coding}.
In this experiment, we vary the local attention window around each token. 

Table~\ref{table-window-in-Longformer} shows that even using a small window size, the micro $F_1$ score on MIMIC-III development set is still close to using a larger window size.
We observe the same pattern on ECtHR and 20 News (See Table~\ref{appendix-table-window-in-Longformer} in the Appendix).
A major advantage of using smaller local attention windows is the faster computation for training and evaluation.

\paragraph{Considering a small number of tokens for global attention improves the stability of the training process.} Longformer relies heavily on the [CLS] token, which is the only token with global attention---attending to all other tokens and all other tokens attending to it.
%The vanilla Longformer for document classification model relies heavily on the [CLS] token: 1) it is the only token with global attention---attending to all other tokens and all other tokens attending to it; and, 2) only the hidden representation of the [CLS] token is taken as input of the final classifier.
We investigate whether allowing more tokens to use global attention can improve model performance, and if yes, how to choose which tokens to use global attention.

\begin{figure}[t]
\begin{subfigure}{\linewidth}
  \centering
  \includegraphics[width=\textwidth]{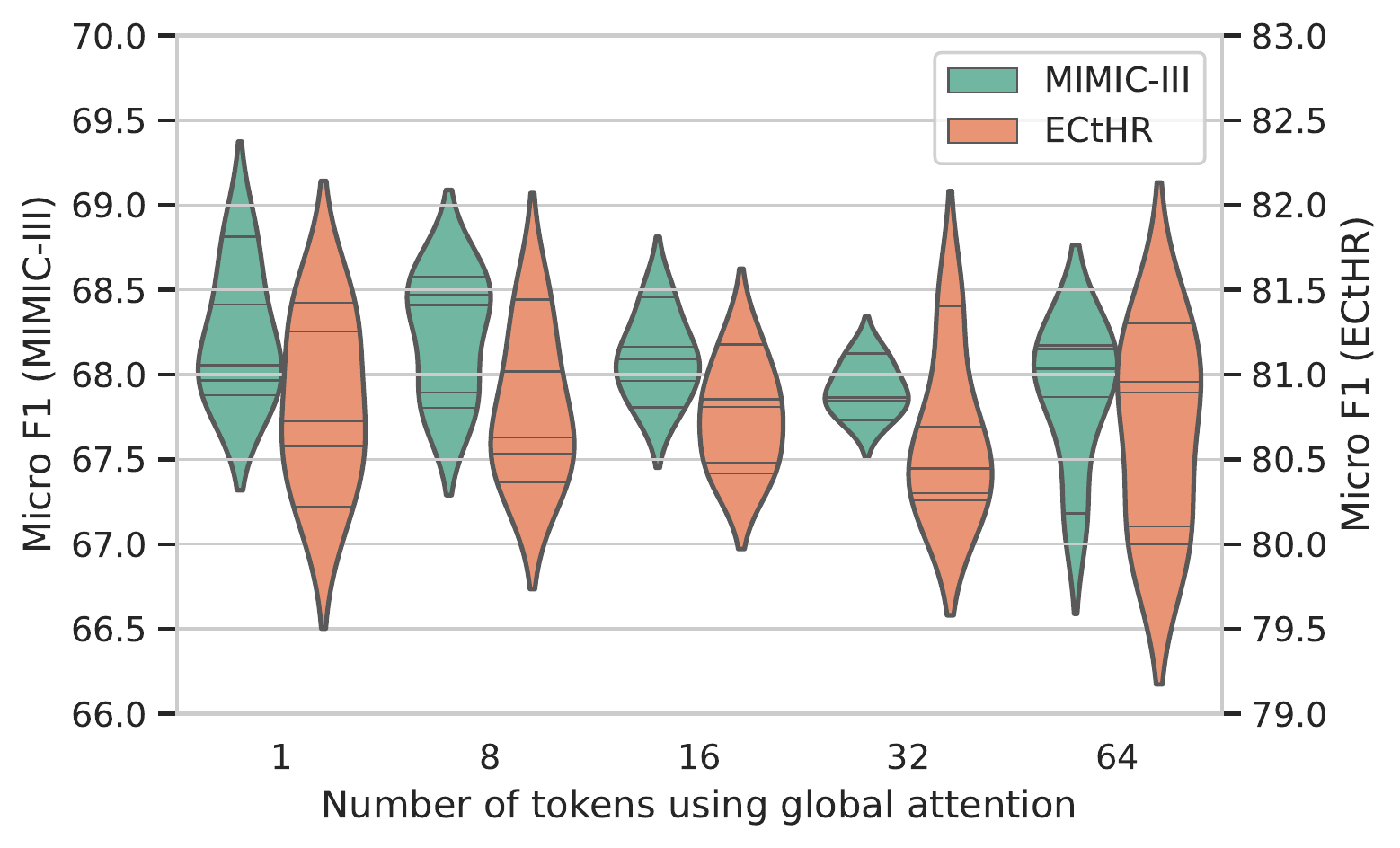}
\end{subfigure}%
%\begin{subfigure}{\linewidth}
%  \centering
%  \includegraphics[width=\textwidth]{images/20211103A-1.pdf}
%\end{subfigure}
\caption{The effect of applying global attention on more tokens, which are evenly chosen based on their positions. In the baseline model (first column), only the [CLS] token uses global attention.}
\label{figure-global-attention-results}
\vspace{-4mm}
\end{figure}

Figure~\ref{figure-global-attention-results} shows that adding more tokens using global attention does not improve $F_1$ score, while a small number of additional global attention tokens can make the training more stable.
%This reflects the \textcolor{red}{position embeddings play an important role than word embeddings?}

\subparagraph{Equally distributing global tokens across the sequence is better than content-based attribution.}
We consider two approaches to choose additional tokens that use global attention: position based or content based. In the position-based approach, we distribute $n$ additional tokens at equal distances. 
For example, if $n=4$ and the sequence length is $4096$, there are global attention on tokens at position $0$, $1024$, $2048$ and $3072$. In the content-based approach, we identify informative tokens, using TF-IDF (Term Frequency–Inverse Document Frequency) within each document, and we apply global attention on the top-$K$ informative tokens, together with the [CLS] token. 
Results show that the position based approach is more effective than content based (see Table~\ref{appendix-table-global-attention-content-based} in the Appendix).

\subsection{Hierarchical Transformers}

\begin{figure}[t]
\begin{subfigure}{0.95\linewidth}
  \centering
  \includegraphics[width=\textwidth]{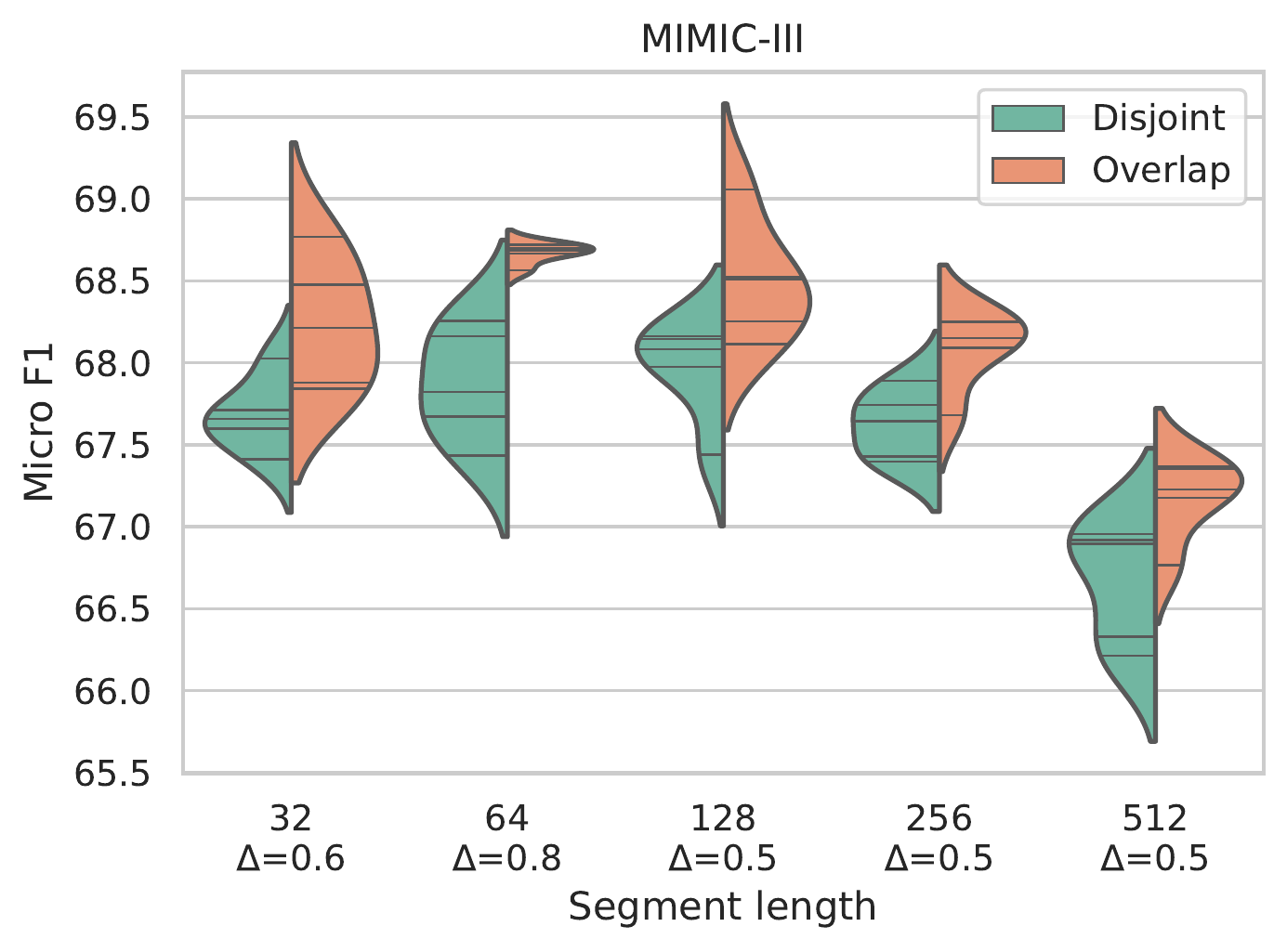}
  \centering
  \includegraphics[width=\textwidth]{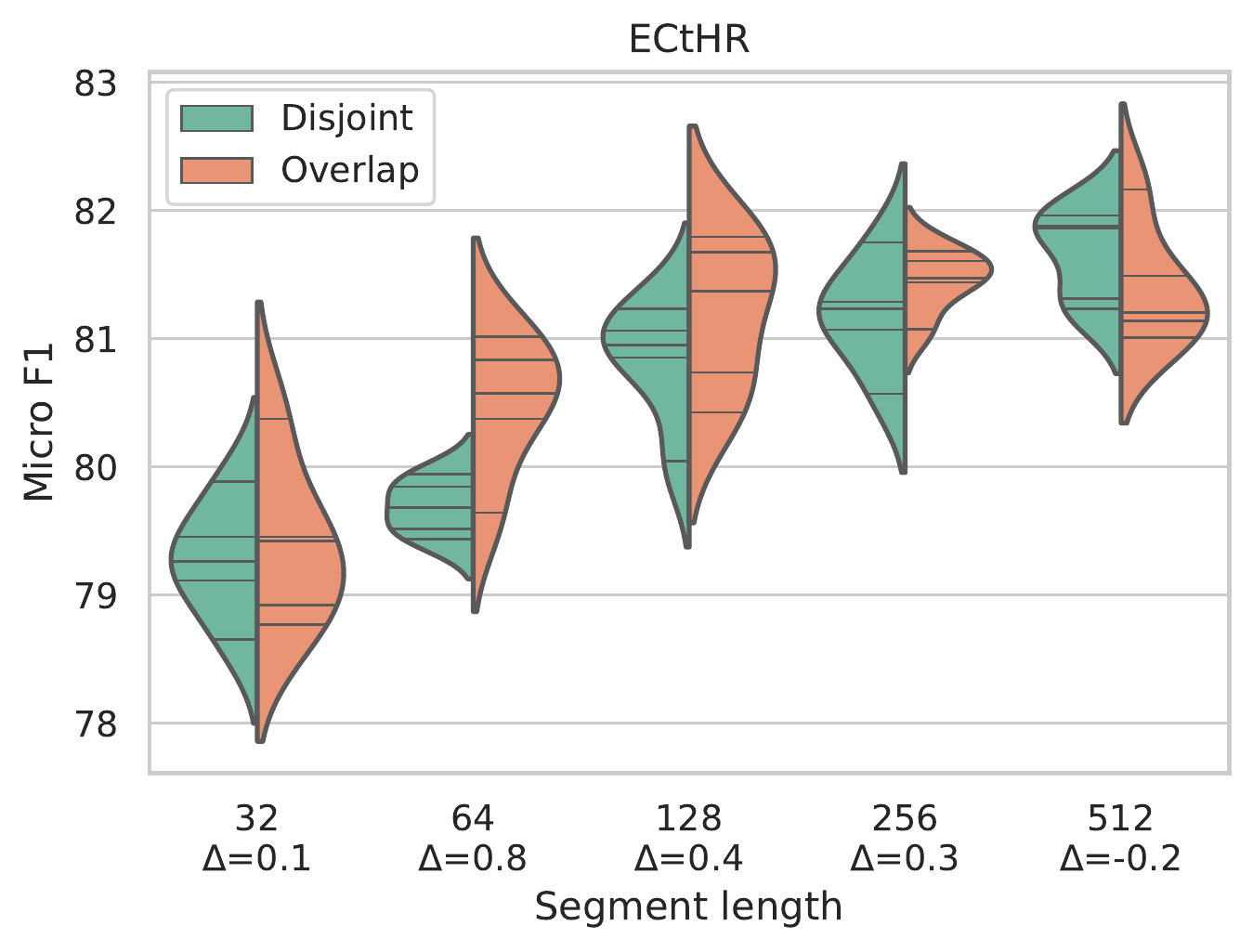}
\end{subfigure}
\vspace{-2mm}
\caption{The effect of varying the segment length and whether allowing segments to overlap in the hierarchical Transformers. $\Delta$: improvement due to overlap.}
\label{figure-hierarchical-segment-length}
\vspace{-4mm}
\end{figure}

\paragraph{The optimal segment length is dataset dependent.}\hspace{-2ex}
\citet{ji-aalto-2021-icd-coding} and \citet{gao-ornl-2021-jbhi-icd-coding} reported negative results with a hierarchical Transformer with a segment length of $512$ tokens on the MIMIC-III dataset. Their methods involved splitting a document into equally sized segments, which were processed using a shared BERT encoder. Instead of splitting the documents into such large segments, we investigate the impact of segment length and preventing context fragmentation.
% Recall that a small local attention window size (e.g., $32$) still achieves close results to the one using large window size (i.e., 512).

Figure~\ref{figure-hierarchical-segment-length} (left side in each violin plot) shows that there is no optimal segment length across both MIMIC-III and ECtHR. Small segment length works well on MIMIC-III, and using segment length greater than $128$ starts to decrease the performance. In contrast, the ECtHR dataset benefits from a model with larger segment lengths.
The optimal performing segment length on 20 News and Hyperpartisan are 256 and 128, respectively (See Table~\ref{appendix-table-segment-length-no-overlap} in the Appendix).

\subparagraph{Splitting documents into overlapping segments can alleviate the context fragmentation problem.}\hspace{-2ex} Splitting a long document into smaller segments may result in the problem of context fragmentation, where a model lacks the information it needs to make a prediction~\cite{dai-cmu-2019-acl-transformer-xl,ding-baidu-2021-acl-ernie-doc}.
Although, the hierarchical model uses a second-order transformer to fuse and contextualise information across segments, we investigate a simple way to alleviate context fragmentation by allowing segments to overlap when we split a document into segments.
That it, except for the first segment, the first $\frac{1}{4}n$ tokens in each segment are taken from the previous segment, where $n$ is the segment length.
Figure~\ref{figure-hierarchical-segment-length} (right side in each violin plot) show that this simple strategy can easily improve the effectiveness of the model.

\paragraph{Splitting based on document structure.}\hspace{-2ex} \citet{chalkidis-jana-2022-acl-lexglue} argue that we should follow the structure of a document when splitting it into segments~\cite{tang-hit-2015-emnlp-document,yang-cmu-2016-naacl-han}.
They propose a hierarchical Transformer for the ECtHR dataset that splits a document at the paragraph level, reading up to $64$ paragraphs of $128$ token each (8192 tokens in total).

% Inspired by their efforts, w
We investigate whether splitting based on document structure is better than splitting a long document into segments of same length.
Similar to their model, we consider each paragraph as a segment and all segments are then truncated or padded to the same segment length.
We follow~\citet{chalkidis-jana-2022-acl-lexglue} and use segment length ($l$) of $128$ on ECtHR, and tune $l\in$\{32, 64, 128\} on MIMIC-III.\footnote{Note that since we need to pad short segments, therefore, a larger maximum sequence length is required to preserve the same information as in evenly splitting.}

Results show that splitting by the paragraph-level document structure 
% only outperforms the naive approach at $8192$ tokens 
does not improve performance on the ECtHR dataset. On MIMIC-III, splitting based on document structure substantially underperforms evenly splitting the document (Figure~\ref{figure-sentence_boundary_vs_evenly_splitting} in the Appendix) .\vspace{1mm}

% There is inconclusive evidence about whether to split based on the structure of a document.\vspace{1mm}

% Exploiting the typical document structure seems to be insignificant.

%\begin{figure}[h]
%\begin{subfigure}{\linewidth}
%  \centering
  %\includegraphics[width=\textwidth]{images/20211106A-mimic-iii.pdf}
  %\centering
  %\includegraphics[width=\textwidth]{images/20211106A-ecthr.pdf}
%  \includegraphics[width=\textwidth]{images/label_wise_attention_vs_max_pooling.pdf}
%\end{subfigure}
%\caption{Using label-wise attention network on top of contextual segment representations bring large improvement on MIMIC-III, but negligible improvements on ECtHR. \textcolor{red}{TO UPDATE: Longformer results}}
%\label{figure-label-wise-attention-vs-max-pooling}
%\end{figure}

\subsection{Label-wise Attention Network}
%\paragraph{Adding a label-wise attention network on top of the contextual segment representations can further improve the effectiveness when multiple labels are assigned to a document}
Recall from Section \ref{sec:approaches} that our models form a single document vector which is used for the final prediction.
That is, in Longformer, we use the hidden states of the [CLS] token; in hierarchical models, we use the max pooling operation to aggregate a list of contextual segment representations into a document vector. The
Label-Wise Attention Network (LWAN)~\cite{mullenbach-gatech-2018-naacl-caml,xiao-bjtu-2019-emnlp-multilabel,chalkidis-aueb-2020-emnlp-multilabel} is an alternative that allows the model to learn distinct document representations for each label. Given a sequence of hidden representations (e.g., contextual token representations in Longformer or contextual segment representations in hierarchical models: $\dmatrix{S} = [\dvector{s}_0, \dvector{s}_1, \cdots, \dvector{s}_m]$), LWAN can allow each label to learn to attend to different positions via:
\begin{align}
    \dvector{a}_\ell &= \text{SoftMax} ( \dmatrix{S}^\top \dvector{u}_\ell) \\
    \dvector{v}_\ell &= \sum_{i=1}^m \dvector{a}_{\ell,i} \dvector{s}_i \\
    \hat{\dvector{y}}_\ell &= \sigma (\dvector{\beta}_\ell^\top \dvector{v}_\ell)
\end{align}
where $\dvector{u}_\ell$ and $\dvector{\beta}_\ell$ are vector parameters for label $\ell$.

\begin{table*}[t]
    \centering
    \setlength{\tabcolsep}{3pt} % Default value: 6pt
    \begin{tabular}{lc cc cc c}
    \toprule
    & & Macro AUC & Micro AUC & Macro $F_1$ & Micro $F_1$ & P@5  \\
    \midrule
    CAML~\cite{mullenbach-gatech-2018-naacl-caml} & $\mathbb{C}$ & 88.4 & 91.6 & 57.6 & 63.3 & 61.8 \\
    %\citet{dong-2021-jbi-icd-coding} & $\mathbb{C}$ & 88.4 & 91.9 & 56.8 & 64.0 & 62.4 \\
    %\citet{cao-chen-2020-acl-icd-coding} & $\mathbb{C}$ & 89.5 & 92.9 & 60.9 & 66.3 & 63.2 \\
    %MultiResCNN\citet{li-yu-2020-aaai-icd-coding} & $\mathbb{C}$ & 89.9 & 92.8 & 60.6 & 67.0 & 64.1 \\
    PubMedBERT~\cite{ji-aalto-2021-icd-coding} & $\mathbb{T}$ & 88.6 & 90.8 & 63.3 & 68.1 & 64.4 \\
    GatedCNN-NCI~\cite{ji-aalto-2021-acl-icd-coding} & $\mathbb{C}$ & 91.5 & 93.8 & 62.9 & 68.6 & 65.3 \\
    %\citet{xie-xiong-2019-cikm}* & $\mathbb{C}$ & 91.4 & 93.6 & 63.8 & 68.4 & 64.4 \\
    % \citet{tasi-huang-2021-naacl} & $\mathbb{R}$ & --- & --- & 65.4 & 70.2 & --- \\
    LAAT~\cite{vu-csiro-2020-ijcai-laat} & $\mathbb{R}$ & 92.5 & 94.6 & 66.6 & 71.5 & 67.5 \\
    MSMN~\cite{yuan-alibaba-2022-acl-msmn} & $\mathbb{R}$ & \textbf{92.8} & \textbf{94.7} & \textbf{68.3} & \textbf{72.5} & \textbf{68.0} \\
    \midrule
    \multicolumn{5}{l}{Baselines processing up to 512 tokens} \\
    \midrule
    First & $\mathbb{T}$ & 83.0 \tiny $\pm$ 0.1 & 86.0 \tiny $\pm$ 0.1 & 47.0 \tiny $\pm$ 0.4 & 56.1 \tiny $\pm$ 0.2 & 55.4 \tiny $\pm$ 0.2 \\ % 2022-03-02-D
    Random & $\mathbb{T}$ & 82.5 \tiny $\pm$ 0.2 & 85.4 \tiny $\pm$ 0.1 & 42.7 \tiny $\pm$ 0.4 & 51.1 \tiny $\pm$ 0.2 & 52.3 \tiny $\pm$ 0.2 \\ % 2022-03-02-B
    Informative & $\mathbb{T}$ & 82.7 \tiny $\pm$ 0.1 & 85.8 \tiny $\pm$ 0.1 & 46.4 \tiny $\pm$ 0.5 & 55.2 \tiny $\pm$ 0.3 & 54.8 \tiny $\pm$ 0.2 \\ % 2022-03-02-C
    \midrule
    \multicolumn{5}{l}{Long document models} \\
    \midrule
    % Longformer (4096 tokens) & $\mathbb{T}$ & 89.9 \tiny $\pm$ 0.1 & 92.4 \tiny $\pm$ 0.1 & 60.3 \tiny $\pm$ 0.4 & 67.9 \tiny $\pm$ 0.3 & 64.8 \tiny $\pm$ 0.1 \\ % 2021-09-08-A
    Longformer (4096 + LWAN) & $\mathbb{T}$ & 90.0 \tiny $\pm$ 0.1 & 92.6 \tiny $\pm$ 0.2 & 60.7 \tiny $\pm$ 0.6 & 68.2 \tiny $\pm$ 0.2 & 64.8 \tiny $\pm$ 0.2 \\ % 2021-11-11-A
    % Hierarchical (4096 tokens) & $\mathbb{T}$ & 89.3 \tiny $\pm$ 0.1 & 91.9 \tiny $\pm$ 0.1 & 61.1 \tiny $\pm$ 0.2 & 67.8 \tiny $\pm$ 0.2 & 64.2 \tiny $\pm$ 0.1 \\ % 2022-03-30-A
    Hierarchical (4096 + LWAN) & $\mathbb{T}$ & 91.1 \tiny $\pm$ 0.1 & 93.6 \tiny $\pm$ 0.0 & 62.9 \tiny $\pm$ 0.1 & 69.5 \tiny $\pm$ 0.1 & 65.7 \tiny $\pm$ 0.2 \\ % 2022-03-30-A
    Hierarchical (4096 + LWAN + L*) & $\mathbb{T}$ & 91.7 \tiny $\pm$ 0.1 & 94.1 \tiny $\pm$ 0.0 & 65.2 \tiny $\pm$ 0.2 & 71.0 \tiny $\pm$ 0.1 & 66.2 \tiny $\pm$ 0.1 \\ % 2022-03-30-A
    % Hierarchical (6144 tokens) & $\mathbb{T}$ & 89.5 \tiny $\pm$ 0.0 & 92.1 \tiny $\pm$ 0.0 & 61.1 \tiny $\pm$ 0.3 & 68.0 \tiny $\pm$ 0.1 & 64.2 \tiny $\pm$ 0.2 \\ % 2022-03-30-A
    % Hierarchical (6144 tokens + LWAN) & $\mathbb{T}$ & 91.3 \tiny $\pm$ 0.1 & 93.7 \tiny $\pm$ 0.1 & 63.5 \tiny $\pm$ 0.5 & 69.9 \tiny $\pm$ 0.3 & 65.7 \tiny $\pm$ 0.1 \\ % 2022-03-30-A
    % Hierarchical (8192 tokens) & $\mathbb{T}$ & 89.5 \tiny $\pm$ 0.1 & 92.1 \tiny $\pm$ 0.0 & 61.0 \tiny $\pm$ 0.5 & 67.8 \tiny $\pm$ 0.3 & 64.3 \tiny $\pm$ 0.2 \\ % 2022-03-30-A
    Hierarchical (8192 + LWAN) & $\mathbb{T}$ & 91.4 \tiny $\pm$ 0.0 & 93.7 \tiny $\pm$ 0.1 & 63.8 \tiny $\pm$ 0.3 & 70.1 \tiny $\pm$ 0.1 & 65.9 \tiny $\pm$ 0.1 \\ % 2022-03-30-A
    %\midrule
    %\multicolumn{5}{l}{Publicly available domain-specific RoBERTa-LARGE~\cite{lewis-fb-2020-clinicalnlp-bio-lm}} \\
    %\midrule
    
	Hierarchical (8192 + LWAN + L*) & $\mathbb{T}$ & 91.9 \tiny $\pm$ 0.2 & 94.1 \tiny $\pm$ 0.2 & 65.5 \tiny $\pm$ 0.7 & 71.1 \tiny $\pm$ 0.4 & 66.4 \tiny $\pm$ 0.3 \\ % 2022-03-30-A
    %\midrule
    %BERT (first 512 tokens) & $\mathbb{T}$ & 81.3 \tiny $\pm$ 0.3 & 85.0 \tiny $\pm$ 0.3 & 41.3 \tiny $\pm$ 1.2 & 52.3 \tiny $\pm$ 0.6 & 53.5 \tiny $\pm$ 0.2 \\ % %2021-11-13-A
    %RoBERTa (first 512 tokens) & $\mathbb{T}$ & 81.0 \tiny $\pm$ 0.2 & 84.8 \tiny $\pm$ 0.2 & 39.8 \tiny $\pm$ 0.7 & 52.4 \tiny $\pm$ 0.3 & 53.2 \tiny $\pm$ 0.2 \\ % 2021-09-17-C
    %RoBERTa (512 random tokens) & $\mathbb{T}$ & 81.0 \tiny $\pm$ 0.2 & 84.8 \tiny $\pm$ 0.2 & 39.8 \tiny $\pm$ 0.7 & 52.4 \tiny $\pm$ 0.3 & 53.2 \tiny $\pm$ 0.2 \\ % 2021-09-17-C
    %RoBERTa (the 512 most informative tokens) & $\mathbb{T}$ & 81.0 \tiny $\pm$ 0.2 & 84.8 \tiny $\pm$ 0.2 & 39.8 \tiny $\pm$ 0.7 & 52.4 \tiny $\pm$ 0.3 & 53.2 \tiny $\pm$ 0.2 \\ % 2021-09-17-C
    %\midrule
    %Hierarchical (5120 tokens) & $\mathbb{T}$ & 89.5 \tiny $\pm$ 0.1 & 92.0 \tiny $\pm$ 0.1 & 61.7 \tiny $\pm$ 0.5 & 68.2 \tiny $\pm$ 0.3 & 64.5 \tiny $\pm$ 0.2 \\ % 2021-10-11-A
    % Hierarchical (6144 tokens) & $\mathbb{T}$ & 89.4 \tiny $\pm$ 0.1 & 92.0 \tiny $\pm$ 0.1 & 61.5 \tiny $\pm$ 0.4 & 68.1 \tiny $\pm$ 0.2 & 64.3 \tiny $\pm$ 0.2 \\ % 2021-10-11-A
    %\midrule
    %\multicolumn{5}{l}{Transformer-based Models with Label-wise Attention Network} \\
    %\midrule
    %Hierarchical (5120 tokens) & $\mathbb{T}$ & 91.2 \tiny $\pm$ 0.1 & 93.6 \tiny $\pm$ 0.1 & 63.8 \tiny $\pm$ 0.5 & 70.2 \tiny $\pm$ 0.3 & 65.9 \tiny $\pm$ 0.2 \\ % 2021-11-06-A
    \bottomrule
    \end{tabular}
    \caption{Comparison of \tldc against state-of-the-art on the MIMIC-III test set. $\mathbb{C}$: CNN-based models; $\mathbb{R}$: RNN-based models; and $\mathbb{T}$: Transformer-based models. Models marked with an asterisk (*) is domain-specific RoBERTa-Large~\cite{lewis-fb-2020-clinicalnlp-bio-lm}, whereas Longformer and other RoBERTa models are task-adaptive pre-trained base versions.}.
    \label{table-mimic-sota}
    \vspace{-7mm}
\end{table*}
% Models marked with an asterisk (*) exploit the label hierarchy, i.e., they use a better classification component, as defined in Section~\ref{sec:approaches}. Results are sorted by Micro $F_1$. 

\begin{table}[t]
    \centering
    \setlength{\tabcolsep}{3pt}
    \begin{tabular}{r ccc}
    \toprule
    & ECtHR & 20 News & Hyper \\
    \midrule
    First (512) & 73.5 \tiny $\pm$ 0.2 & 86.1 \tiny $\pm$ 0.3 & 92.9 \tiny $\pm$ 3.2 \\ %% 2022-03-02-D
    Random (512) & 79.0 \tiny $\pm$ 0.6 & 85.3 \tiny $\pm$ 0.4 & 88.9 \tiny $\pm$ 2.5 \\ %% 2022-03-02-B
    Informative (512) & 72.4 \tiny $\pm$ 0.2 & 86.2 \tiny $\pm$ 0.3 & 91.7 \tiny $\pm$ 3.2 \\ %% 2022-03-02-C
    Longformer (4096) & 81.0 \tiny $\pm$ 0.5 & \bf 86.3 \tiny $\pm$ 0.5 & \bf 97.9 \tiny $\pm$ 0.7 \\ %% 2022-02-07-A
    Hierarchical (4096) & \bf 81.1 \tiny $\pm$ 0.2 & \bf 86.3 \tiny $\pm$ 0.2 & 95.4 \tiny $\pm$ 1.3 \\ %% 2022-03-12-B
    \bottomrule
    \end{tabular}
    \caption{Comparison of \tldc against baselines processing up to 512 tokens. We report Micro $F_1$ on ECtHR, Accuracy on 20 News and Hyperpartisan datasets.}.
    \label{table-results-ecthr-20news-hyper}
\end{table}

Results show that adding a LWAN improves performance on MIMIC-III (Micro $F_1$ score of $1.1$ with Longformer; $1.8$ with hierarchical models), where on average each document is assigned $6$ labels out of 50 available labels (classes). There is a smaller improvement on ECtHR ($0.4$ with Longformer; $0.1$ with hierarchical models), where the average number of labels per document is $1.5$ out of 10 labels (classes) in total (Table~\ref{appendix-table-hierarchical-lwan} in the Appendix).

%We also investigate a variant that builds the label-wise attention network on top of non-contextual segment representation (i.e., $\dmatrix{S}$ = [$\dvector{s}_{0}$, $\dvector{s}_{1}$, $\cdots$, $\dvector{s}_{m}$] in Figure~\ref{figure-hierarchical-transformer}).
%We find it underperforms even max pooling contextual segment representations (\textcolor{red}{results can be found in Appendix~\ref{table-label-wise-attention-vs-transformer}}), demonstrating that the interactions between capturing interactions between segments is a imperative step in hierarchical transformers.

\begin{table}[ht]
    \centering
    \setlength{\tabcolsep}{14pt} % Default value: 6pt
    \begin{tabular}{r cc}
    \toprule
    & Longformer & Hierarchical \\
    Size & (148.6M) & (139.0M) \\
    \midrule
    \multicolumn{3}{c}{Maximum sequence length: 1024} \\
    \midrule
64 & 4.8G & 3.6G \\
128 & 5.0G & 3.8G \\
256 & 5.5G & 4.1G \\
512 & 6.6G & 4.7G \\
    \midrule
    \multicolumn{3}{c}{Maximum sequence length: 4096} \\
    \midrule
64 & 11.8G & 7.8G \\
128 & 12.8G & 8.4G \\
256 & 14.9G & 9.6G \\
512 & 19.4G & 12.2G \\
    \bottomrule
    \end{tabular}
    \caption{A comparison between Longformer and Hierarchical models regarding their GPU memory consumption. The number of parameters are listed in the table header. Size refers to the local attention window size in Longformer and the segment length in hierarchical method, respectively.}
    \label{appendix-table-compare-longformer-hierarchical}
\end{table}

\subsection{Comparison with State of the art}
\label{section-compare-to-sota}

We compare \tldc models against recently published results on MIMIC-III, as well as baseline models that process up to 512 tokens.
In addition to the common practice of truncating long documents (i.e., using the first 512 tokens), we consider two alternatives that either randomly choose 512 tokens from the document as input or take as input the most informative 512 tokens, identified using TF-IDF scores.

Results in Table~\ref{table-mimic-sota} and~\ref{table-results-ecthr-20news-hyper} show that there is a clear benefit from being able to process longer text.
Both the Longformer and hierarchical Transformers outperform baselines that process up to 512 tokens with a large margin on MIMIC-III and ECtHR, whereas relatively small improvements on 20 News and Hyperpartisan.
It is also worthy noting that, among these baselines, there is no single best strategy to choose which 512 tokens to process.
Using the first 512 tokens works well on MIMIC-III and Hyperpartisan datasets, but it performs much worse than 512 random tokens on ECtHR.

Finally, Longformer, which can process up to $4096$ tokens, achieves competitive results with the best performing CNN-based model~\cite{ji-aalto-2021-acl-icd-coding} on MIMIC-III. 
By processing longer text and using the RoBERTa-Large model, the hierarchical models further improve the performance, leading to comparable results of RNN-based models~\cite{vu-csiro-2020-ijcai-laat,yuan-alibaba-2022-acl-msmn}.
We hypothesise that further improvements can be observed when \tldc models are enhanced with better hierarchy-aware classifier as in~\citet{vu-csiro-2020-ijcai-laat} or code synonyms are used for training as in~\citet{yuan-alibaba-2022-acl-msmn}.

%a similar hierarchy-aware classifier could lead to comparable or even better results.
% Note that~\citeauthor{xie-xiong-2019-cikm} and \citet{vu-csiro-2020-ijcai-laat} truncate all documents to a maximum sequence length of $4000$ words ($\approx$ $4,932$ subtokens, see Appendix Table~\ref{appendix-table-vanilla-roberta}).
% all CNN-based models by $1.8$ points. Our Transformer-based models only underperform the RNN-based model, which additionally exploits the label hierarchy of ICD codes~\cite{vu-csiro-2020-ijcai-laat}. We hypothesize that using a similar hierarchy-aware classifier could lead to comparable or even better results.

%\begin{itemize}
%    \item Our results are competitive with recently published results
%    \item \cite{xie-xiong-2019-cikm,vu-csiro-2020-ijcai-laat} take the hierarchical relationships among medical codes into consideration and achieve higher $F_1$, which can be our future work
%\end{itemize}

%The ECtHR dataset~\cite{chalkidis-jana-2022-acl-lexglue} is a very recently released dataset, where the authors used hierarchical Transformers.
%Our results are on par with their results (See Appendix Table~\ref{appendix-table-ecthr-results}).

\subsection{Comparison in terms of of GPU memory consumption}
GPU memory becomes a big constraint when Transformer-based models are trained on long text.
Table~\ref{appendix-table-compare-longformer-hierarchical} shows a comparison between Longformer and Hierarchical models regarding the number of parameters and their GPU consumption.
We use batch size of 2 in these experiments, and measure the impact of attention window size and segment length on the memory footprint.
We find that Hierarchical models require less GPU memory than Longformer in general, and it is possible to set smaller local window size in Longformer or segment length in hierarchical models to fit the model using smaller GPU memory.
Recall that small local attention windows are effective in Longformer, and the optimal segment length in hierarchical models is dataset dependent.
%We set the maximum sequence length as $4096$ and use $128$ for both local window size in Longformer and segment length in hierarchical models.
%Note that in this experiment we try to make full use of GPU memory ($24$G) via setting as large as possible batch size (i.e., training batch size of $5$ and test batch size of $256$ in Longformer; $7$ and $256$ in hierarchical model).

\section{Practical Advice} 
\label{section-practical-advice}
We compile several questions that practitioners may ask regarding long document classification and provide answers based on our results:

\paragraph{Q1} When should I start to consider using long document classification models?
\paragraph{A} We suggest using \tldc models if you work with datasets consisting of long documents (e.g., 2K tokens on average).
We notice that on 20 News dataset, the gap between baselines that process 512 tokens and long document models is negligible.\footnote{Although Hyperpartisan is a widely used benchmark for long document models, we do not recommend drawing practical conclusions based on our results because we observe high variance when we run experiments using different GPUs or CUDA versions. 
We attribute this may to the small size (65) of its test set and the subjectivity of the task.}

\paragraph{Q2} Which model should I choose? Longformer or hierarchical Transformers?
\paragraph{A} We suggest Longformer as the starting point if you do not plan on extensively tuning hyperparameters.
We find the default config of Longformer is robust, although it is possible to set a moderate size (64-128) of local attention window to improve efficiency without sacrificing effectiveness, and a small number of additional global attention tokens to make the training more stable.
On the other hand, hierarchical Transformers may benefit from careful hyperparameter tuning (e.g., document splitting strategy, using LWAN).
We suggest splitting a document into small non-structure-derived segments (e.g., $128$ tokens) which overlap as a starting point when employing hierarchical Transformers.

We also note that the publicly available Longformer models can process sequences up-to 4096 tokens, whereas hierarchical Transformers can be easily extended to process much longer sequence.

%\noindent\textbf{Take-Away \#1:} We suggest task-adaptive pre-training as a first step when applying pre-trained models on domain-specific datasets, as it is effective and cheap.
%The following experiments on MIMIC-III and ECtHR are based on task-adaptive pre-trained Longformer and RoBERTa models.
%Therefore, we suggest a moderate size (64-128) of local attention window. We use a local window of $128$ in the following experiments.
% TODO: COMPARE RoBERTa-ECtHR AGAINST LEGAL-BERT
%\noindent\textbf{Take-Away \#2:} We suggest the following hyperparameters for Longformer for long-document classification: a local attention window of 128 tokens, and 16 equally-distributed global attention tokens.
%\noindent\textbf{Take-Away \#3:} We suggest splitting a document into small non-structure-derived segments (e.g., $128$) which overlap as a starting point when employing hierarchical Transformers. 

\section{Related Work}

\paragraph{Long document classification}
Document length was not a point of controversy in the pre-neural era of NLP, where documents are encoded with Bag-of-Word representations, e.g., TF-IDF scores. The issue arised with the introduction of deep neural networks. 
\citet{tang-hit-2015-emnlp-document} use CNN and BiLSTM based hierarchical networks in a bottom-up fashion, i.e., first encode sentences into vectors, then combine those vectors in a single document vector.
% learn vector-based document representation for sentiment analysis in a bottom-up fashion: use a CNN or LSTM to learn sentence vectors and then a bi-directional gated recurrent network to learn a document vector.
Similarly, \citet{yang-cmu-2016-naacl-han} incorporate the attention mechanism when constructing the sentence and document representation.
Hierarchical variants of BERT have also been explored for document classification~\cite{mulyar-jhu-2019-document-classification,chalkidis-jana-2022-acl-lexglue}, abstractive summarization~\cite{zhang-microsoft-2019-acl-hierarchical}, semantic matching~\cite{yang-google-2020-cikm-smith}. Both \citeauthor{zhang-microsoft-2019-acl-hierarchical}, and \citeauthor{yang-google-2020-cikm-smith} also propose specialised pre-training tasks to explicitly capture sentence relations within a document.
A very recent work by~\citet{park-amazon-2022-acl-document-classification} shows that \tldc do not perform consistently well across datasets that consist of $700$ tokens on average.

Methods of modifying transformer architecture for long documents can be categorised into two approaches: \emph{recurrent} Transformers and \emph{sparse} attention Transformers.
The recurrent approach processes segments moving from left-to-right~\cite{dai-cmu-2019-acl-transformer-xl}.
To capture bidirectional context, \citet{ding-baidu-2021-acl-ernie-doc} propose a retrospective mechanism in which segments from a document are fed twice as input.
Sparse attention Transformers have been explored to reduce the complexity of self-attention, via using dilated sliding window~\cite{child-openai-2019-sparse-transformers}, and locality-sensitive hashing attention~\cite{kitaev-google-2020-iclr-reformer}.
Recently, the combination of local (window) and global attention are proposed by~\citet{beltagy-allenai-2020-longformer} and~\citet{zaheer-google-2020-neurips-bigbird}, which we have detailed in Section~\ref{sec:approaches}.

\paragraph{ICD Coding} The task of assigning most relevant ICD codes to a document, e.g., radiology report~\cite{pestian-brew-2007-bionlp-shared-task}, death certificate~\cite{koopman-csiro-2015-icd-coding} or discharge summary~\cite{johnson-mit-2016-mimic-iii}, as a whole, has a long history of development~\cite{farkas-szarvas-2008-icd-coding}.
Most existing methods simplified this task as a text classification problem and built classifiers using CNNs~\cite{karimi-csiro-2017-bionlp-icd-coding} or LSTMs~\cite{xie-petuum-2018-acl-icd-coding}.
%Since the number of unique ICD codes is very large, methods are proposed to exploit relation between codes based on label co-occurrence~\cite{dong-ed-2021-jbi-icd-coding}, label count~\cite{du-nih-2019-jamia-ml-net}, label hierarchical~\cite{vu-csiro-2020-ijcai-laat}, knowledge graph~\cite{xie-fudan-2019-cikm-icd-coding,cao-ac-2020-acl-icd-coding,lu-monash-2020-emnlp-label-graph}, code's textual descriptions~\cite{mullenbach-gatech-2018-naacl-caml,xie-petuum-2018-acl-icd-coding,rios-kavuluru-2018-emnlp-multilabel}. More recently, \citet{ji-aalto-2021-icd-coding,gao-ornl-2021-jbhi-icd-coding} investigate various methods of applying BERT on ICD coding.
Since the number of unique ICD codes is very large, methods are proposed to exploit relation between codes based on label co-occurrence~\cite{dong-ed-2021-jbi-icd-coding}, label count~\cite{du-nih-2019-jamia-ml-net}, knowledge graph~\cite{xie-fudan-2019-cikm-icd-coding,cao-ac-2020-acl-icd-coding,lu-monash-2020-emnlp-label-graph}, code's textual descriptions~\cite{mullenbach-gatech-2018-naacl-caml,rios-kavuluru-2018-emnlp-multilabel}. More recently, \citet{ji-aalto-2021-icd-coding,gao-ornl-2021-jbhi-icd-coding} investigate various methods of applying BERT on ICD coding.
Different from our work, they mainly focus on comparing domain-specific BERT models that are pre-trained on various types of corpora.
\citeauthor{ji-aalto-2021-icd-coding} show that PubMedBERT---pre-trained from scratch on PubMed abstracts---outperforms other variants pre-trained on clinical notes or health-related posts; \citeauthor{gao-ornl-2021-jbhi-icd-coding} show that BlueBERT---pre-trained on PubMed and clinical notes---performs best.
However, both report that Transformers-based models perform worse than CNN-based ones.

\section{Conclusions}
Transformers have previously been criticised for being incapable of long document classification. In this paper, we carefully study the role of different components of Transformer-based long document classification models.
By conducting experiments on MIMIC-III and other three datasets (i.e., ECtHR, 20 News and Hyperpartisan), we observe clear improvements in performance when a model is able to process more text. Firstly, Longformer, a sparse attention model, which can process up to $4096$ tokens, achieves competitive results with CNN-based models on MIMIC-III; its performance is relatively robust; a moderate size of local attention window (e.g., 128) and a small number (e.g., 16) of evenly chosen tokens with global attention can improve the efficiency and stability without sacrificing its effectiveness. Secondly, hierarchical Transformers outperform all CNN-based models by a large margin; the key design choice is how to split a document into segments which can be encoded by pre-trained models; although the best performing segment length is dataset dependent, we find splitting a document into small overlapping segments (e.g., 128 tokens) is an effective strategy. Taken together, these experiments rebut the criticisms of Transformers for long document classification.

\iffalse

\section{Limitations}
Long document classification datasets are usually annotated using a large number of labels. 
For example, the complete MIMIC-III dataset contains XXX unique labels, and the ECtHR dataset contains XXX unique labels. 
As we mentioned in Section 2, we focus on building document representation and leave the challenge of learning with a \emph{large target label} set for future work.
Therefore, in this paper, we follow previous work~\cite{mullenbach-gatech-2018-naacl-caml,chalkidis-jana-2022-acl-lexglue} and consider a subset of frequent labels only. That is, we use the top 50 frequent labels in MIMIC-III and 10 labels in ECtHR.

\fi

\section*{Acknowledgments}
This work is funded by the Innovation Fund Denmark under the AI4Xray project.
Xiang Dai is funded by CSIRO Precision Health Future Science Platform.
Ilias Chalkidis is funded by the Innovation Fund Denmark under File No. 0175-00011A.
This project was also undertaken with the assistance of resources and services from the National Computational Infrastructure (NCI), which is supported by the Australian Government.

%% file: appendix.tex
\section{Appendix}

\subsection{Limitations}
Long document classification datasets are usually annotated using a large number of labels. 
For example, the complete MIMIC-III dataset contains $8,692$ unique labels. 
As we mentioned in Section 2, we focus on building document representation and leave the challenge of learning with a \emph{large target label} set for future work.
Therefore, in this paper, we follow previous work~\cite{mullenbach-gatech-2018-naacl-caml,chalkidis-jana-2022-acl-lexglue} and consider a subset of frequent labels in MIMIC-III and ECtHR. 

\subsection{Dataset statistics}

Table~\ref{table-data-statistics} shows the descriptive statistics of four datasets we use.

\begin{table}[ht]
    \centering
    \begin{tabular}{r rrr}
    \toprule
    & Train & Dev & Test\\
    \midrule
    \bf MIMIC-III \\
Documents & 8,066 & 1,573 & 1,729 \\
Unique labels & 50 & 50 & 50 \\
Avg. tokens & 2,260 & 2,693 & 2,737 \\
    \midrule
    \bf ECtHR \\
Documents & 8,866 & 973 & 986 \\
Unique labels & 10 & 10 & 10 \\
Avg. tokens & 2,140 & 2,345 & 2,532 \\
    \midrule
    \bf Hyperpartisan \\
Documents & 516 & 64 & 65 \\
Unique labels & 2 & 2 & 2 \\
Avg. tokens & 741 & 707 & 845 \\
    \midrule
    \bf 20 News \\
Documents & 10,183 & 1,131 & 7,532 \\
Unique labels & 20 & 20 & 20 \\
Avg. tokens & 613 & 627 & 551 \\
    \bottomrule
    \end{tabular}
    \caption{Statistics of the datasets. The number of tokens is calculated using RoBERTa tokenizer.}
    \label{table-data-statistics}
    \vspace{-4mm}
\end{table}

\subsection{Details of task-adaptive pre-training}
\label{appendix-section-pretraining-details}

Hyperparameters and training time for task-adaptive pre-training can be found in Table~\ref{table-hyperparameter-tapt}.
% Note that the main consideration of choosing these hyperparameters is the training budget.

\begin{table}[ht]
    \centering
    \begin{tabular}{r cc c cc c}
    \toprule
    & Longformer & RoBERTa \\ 
    \midrule
    Max sequence & 4096 & 128 \\
    Batch size & 8 & 128 \\
    Learning rate & 5e-5 & 5e-5 \\
    Training epochs & 6 & 15 \\
    Training time & \multirow{2}{*}{$\approx$ 130}  & \multirow{2}{*}{$\approx$ 40} \\
    (GPU-hours) & & \\
    \bottomrule
    \end{tabular}
    \caption{Hyperparameters and training time (measured on MIMIC-III dataset) for task-adaptive pre-training Longformer and RoBERTa. Batch size = batch size per GPU $\times$ num. GPUs $\times$ gradient accumulation steps.}
    \label{table-hyperparameter-tapt}
\end{table}

\subsection{Details of classification experiments}
\paragraph{Preprocessing}
We mainly follow~\citet{mullenbach-gatech-2018-naacl-caml} to preprocess the MIMIC-III dataset.
That is, we lowercase the text, remove all punctuation marks and tokenize text by white spaces.
The only change we make is that we normalise numeric (e.g., convert `2021` to `0000`) instead of deleting numeric-only tokens in~\citet{mullenbach-gatech-2018-naacl-caml}.
We did not apply additional preprocessing to ECtHR and 20 News.
We follow~\citet{beltagy-allenai-2020-longformer} to preprocess the Hyperpartisan dataset.\footnote{\href{https://github.com/allenai/longformer/blob/master/scripts/hp_preprocess.py}{https://github.com/allenai/longformer}}

\paragraph{Training}
We fine-tune the multilabel classification model using a binary cross entropy loss.
That is, given an training example whose ground truth and predicted probability for the $i$-th label are $y_i$ (0 or 1) and $\hat{y}_i$, we calculate its loss, over the $C$ unique classification labels, as:
\[
\mathcal{L} = \sum_{i=1}^C -y_i \log (\hat{y}_i) - (1-y_i) \log (1 - \hat{y}_i).
\]

For the multiclass and binary classification tasks, we fine-tune using the cross entropy loss, where $\hat{y}_g$ is the predicted probability for the gold label:
\[
\mathcal{L} = - \log (\hat{y}_g),
\]

We use the same effective batch size ($16$), learning rate ($2$e-$5$), maximum number of training epochs ($30$) with early stop patience ($5$) in all experiments.
We also follow Longformer~\cite{beltagy-allenai-2020-longformer} and set the maximum sequence length as $4096$ in most of the experiments unless other specified.
We fine-tune all classification models on Quadro RTX 6000 ($24$ GB GPU memory) or Tesla V100 ($32$ GB GPU memory).
If one batch of data is too large to fit into the GPU memory, we use gradient accumulation so that the effective batch sizes (batch size per GPU $\times$ gradient accumulation steps) are still the same.

We repeat all experiments five times with different random seeds. 
The model which is most effective on the development set, measured using the micro $F_1$ score (multilabel) or accuracy (multiclass and binary), is used for the final evaluation.

\subsection{A comparison between clinical notes and legal cases}
\label{appendix-section-clinical-note-and-legal-document}
Although we usually use the term \textit{domain} to indicate that texts talk about a narrow set of related concepts (e.g., clinical concepts or legal concepts), text can vary along different dimensions~\cite{ramponi-plank-2020-coling-uda}.
%In this work, we use MIMIC-III and ECtHR which are sourced from clinical notes and legal cases.
%In addition to the statistics difference which we show in Table~\ref{table-data-statistics}, there are two other differences worthy considering:

In addition to the statistics difference between MIMIC-III and ECtHR, which we show in Table~\ref{table-data-statistics}, there is another difference worthy considering:
clinical notes are private as they contain protected health information. 
Even those clinical notes after de-identification are usually not publicly available (e.g., downloadable using web crawler).
In contrast, legal cases have generally been allowed and encouraged to share with the public, and thus become a large portion of crawled pre-training data~ \cite{dodge-allenai-2021-emnlp-corpora}. 
\citeauthor{dodge-allenai-2021-emnlp-corpora} find that legal documents, especially U.S. case law, are a significant part of the C4 corpus, a cleansed version of CommonCrawl used to pre-train RoBERTa models. The ECtHR proceedings are also publicly available via HUDOC, the court's database.

We suspect task-adaptive pre-training is more useful on MIMIC-III than on ECtHR (Figure~\ref{figure-tapt-results}) may relate to this difference.
Therefore, we evaluate the vanilla RoBERTa on MIMIC-III and ECtHR regarding tokenization and language modelling.
A comparison of the fragmentation ratio using the tokenizer and perplexity using the language model can be found in Table~\ref{appendix-table-vanilla-roberta}.

%\item Clinical notes are usually written by practitioners under time pressure and the main purpose of writing is for \emph{efficient} communication.
    %Therefore, clinical notes are usually syntactically noisy and contain a lot of abbreviations and jargon.
    %In contrast, legal cases are usually the outcome of collaborative writing. 
    %That is, a group of people may edit the text several times to make the text to be more comprehensible.
    %Therefore, legal cases are usually well organised and professionally edited.

\begin{table}[ht]
    \centering
    \setlength{\tabcolsep}{4pt} % Default value: 6pt
    \begin{tabular}{r cc}
    \toprule
     & MIMIC-III & ECtHR \\ 
    \midrule
    Fragmentation ratio & 1.233 & 1.118 \\
    Perplexity & 1.351 & 1.079 \\
    \bottomrule
    \end{tabular}
    \caption{Evaluating vanilla RoBERTa on MIMIC-III and ECtHR. Lower fragmentation ratio and perplexity indicate that the test data have a higher similarity with the RoBERTa pre-training data.}
    \label{appendix-table-vanilla-roberta}
\end{table}

\subsection{A comparison between TAPT and public available RoBERTa by~\cite{lewis-fb-2020-clinicalnlp-bio-lm}}

\begin{figure}[b]
    \centering
    \includegraphics[width=\linewidth]{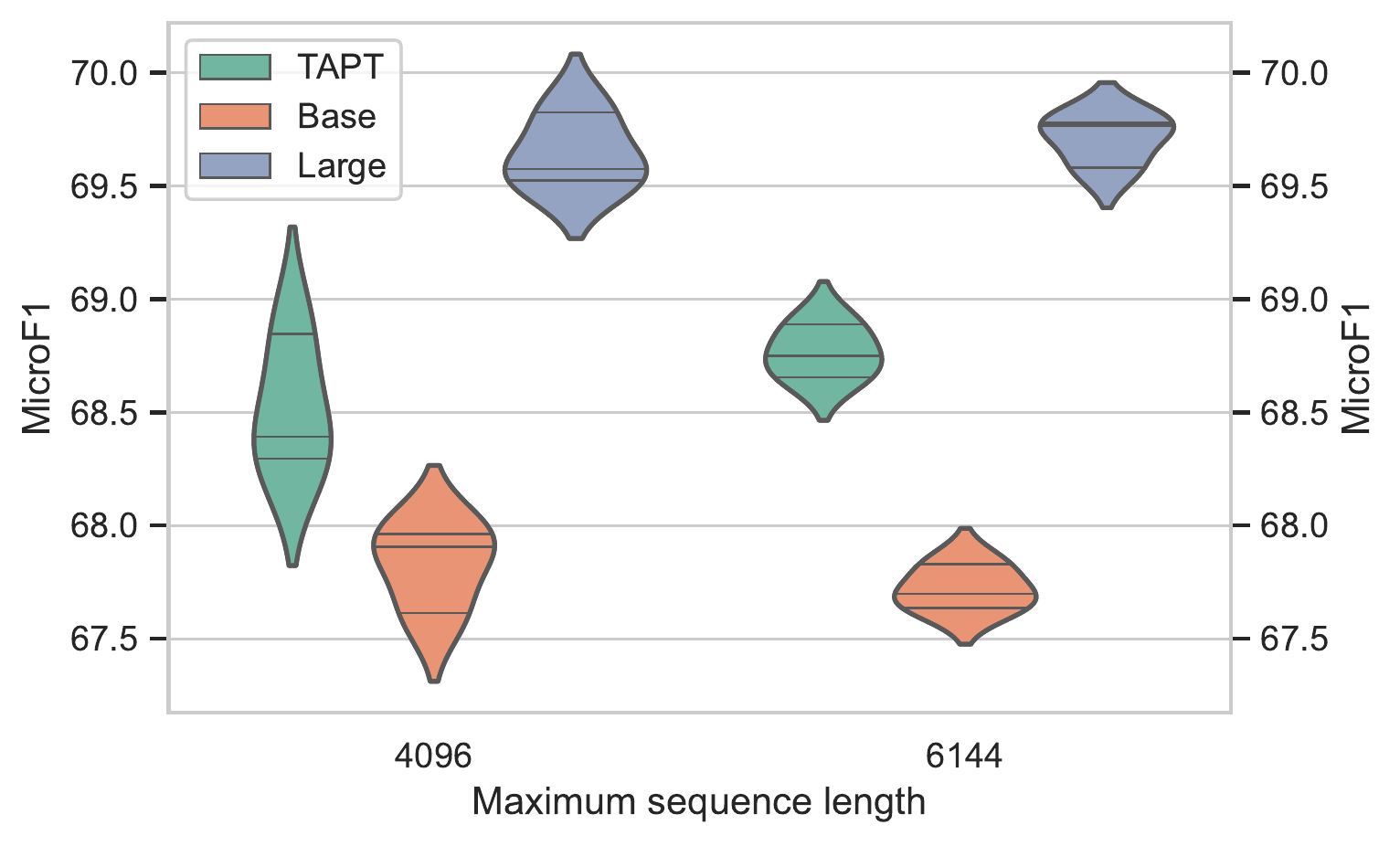}
    \caption{A comparison of task-adaptive pre-trained RoBERTa against public available domain-specific RoBERTa. Both Base and Large RoBERTa models are trained from scratch on biomedical articles and clinical notes~\cite{lewis-fb-2020-clinicalnlp-bio-lm}.}
    \label{appendix-figure-compare-roberta-version}
\end{figure}

We compare our TAPT-RoBERTa against publicly available domain-specific RoBERTa~\cite{lewis-fb-2020-clinicalnlp-bio-lm}, which are trained from scratch on biomedical articles and clinical notes, in hierarchical models.
In these experiments, we split long documents into overlapping segments of 64 tokens.
Results in Figure~\ref{appendix-figure-compare-roberta-version} show that TAPT-RoBERTa outperforms domain-specific base model, but underperforms the larger model.

\subsection{Results on ECtHR test set}

Results in Table~\ref{appendix-table-ecthr-results} show that our results are higher than the ones reported in~\cite{chalkidis-jana-2022-acl-lexglue}.
\citeauthor{chalkidis-jana-2022-acl-lexglue} compare different BERT variants including domain-specific models, whereas we use task-adaptive pre-trained models.
Regarding hierarchical method, we split a document into overlapping segments, each of which has $512$ tokens. We use the default setting for Longformer as in~\citet{beltagy-allenai-2020-longformer}. 

\begin{table}[ht]
    \centering
    %\small
    %\setlength{\tabcolsep}{2pt} % Default value: 6pt
    \begin{tabular}{r cc}
    \toprule
    & Macro $F_1$ & Micro $F_1$\\
    \midrule
    RoBERTa & 68.9 & 77.3 \\
    CaseLaw-BERT & 70.3 & 78.8 \\
    BigBird & 70.9 & 78.8 \\
    DeBERTa & 71.0 & 78.8 \\
    Longformer & 71.7 & 79.4 \\
    BERT & 73.4 & 79.7 \\
    Legal-BERT & 74.7 & 80.4 \\
    \midrule
    Longformer (4096) & 76.0 \tiny $\pm$ 1.4 & 80.7 \tiny $\pm$ 0.3 \\ % 2022-02-07-A
    Hierarchical (4096) & 76.6 \tiny $\pm$ 0.7 & 81.0 \tiny $\pm$ 0.3 \\ % 2022-03-12-B
    \bottomrule
    \end{tabular}
    \caption{Comparison of our results against the results reported in~\cite{chalkidis-jana-2022-acl-lexglue} on the ECtHR test set. Results are sorted by Micro $F_1$.}
    \label{appendix-table-ecthr-results}
\end{table}

\subsection{A comparison between evenly splitting and splitting based on document structure}

\begin{figure}[t]
\begin{subfigure}{\linewidth}
  \centering
  %\includegraphics[width=\textwidth]{images/20211106A-mimic-iii.pdf}
  %\centering
  %\includegraphics[width=\textwidth]{images/20211106A-ecthr.pdf}
  \includegraphics[width=\textwidth]{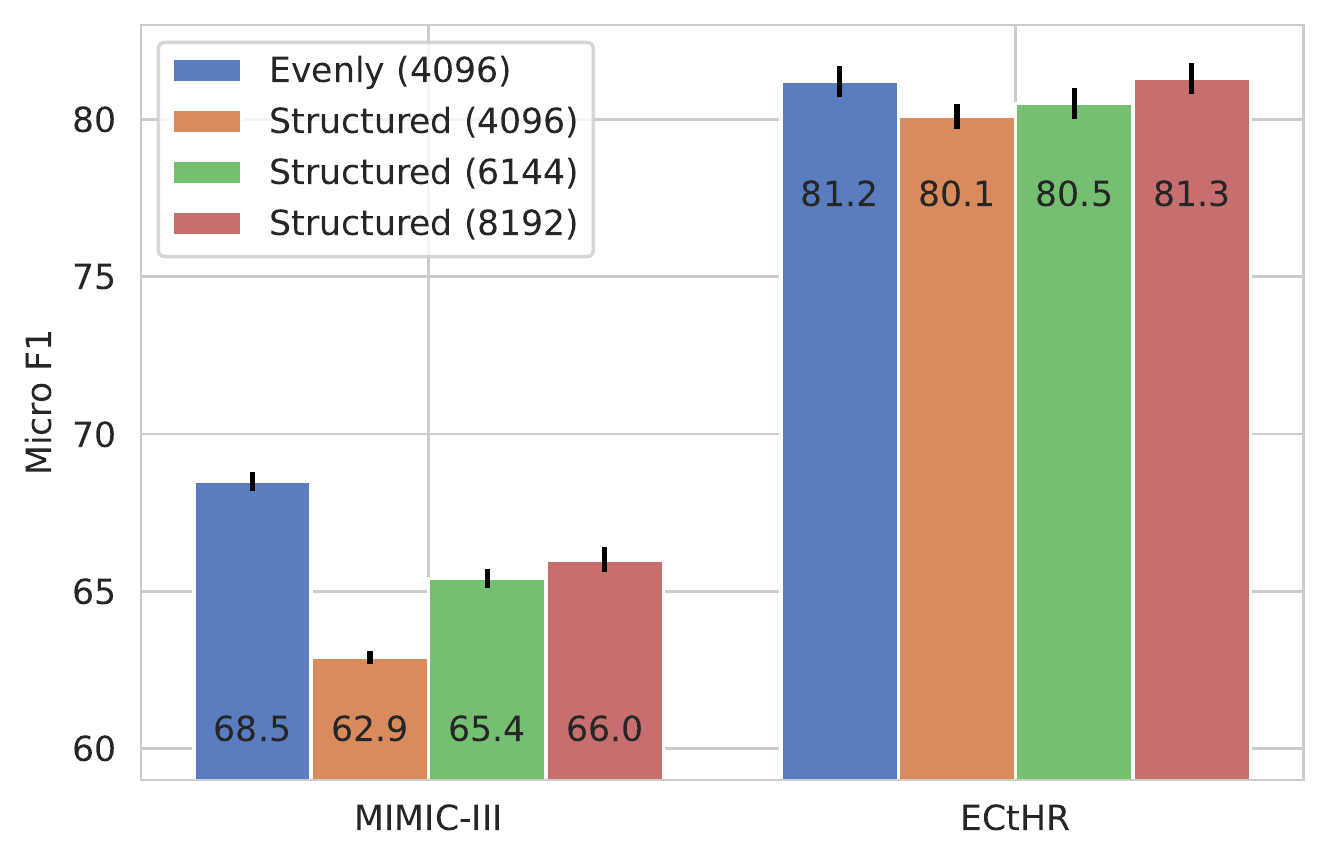}
\end{subfigure}
\vspace{-3mm}
\caption{A comparison between evenly splitting and splitting based on document structure.}
\label{figure-sentence_boundary_vs_evenly_splitting}
\vspace{-5mm}
\end{figure}

Figure~\ref{figure-sentence_boundary_vs_evenly_splitting} shows that splitting by the paragraph level document structure does not improve performance on the ECtHR dataset. On MIMIC-III, splitting based on document structure substantially under-performs evenly splitting the document.

\newpage

\subsection{Detailed results on the development sets}

For the sake of brevity, we use only micro $F_1$ score in most of our illustrations, and we detail results of other metrics in this section.

%Detailed results of Figure\ref{figure-section1_seq_vs_f1},~\ref{figure-tapt-results}, Table~\ref{table-window-in-Longformer}, Figure~\ref{figure-global-attention-results} can be found in Table~\ref{table-detailed-figure-1},~\ref{table-detailed-figure-2},~\ref{appendix-table-window-in-Longformer},~\ref{appendix-table-global-attention} respectively.

%\balance
\begin{table*}[t]
    \centering
    %\small
    %\setlength{\tabcolsep}{2pt} % Default value: 6pt
    \begin{tabular}{rc cc c cc cc}
    \toprule
    && \multicolumn{2}{c}{AUC} & & \multicolumn{2}{c}{$F_1$} & \\
    \cmidrule{3-4} \cmidrule{6-7}
    Max sequence length & & Macro & Micro & & Macro & Micro & & P@5 \\
    \midrule
512 & & 81.4 \tiny $\pm$ 0.1 & 85.1 \tiny $\pm$ 0.2 &  & 39.2 \tiny $\pm$ 0.9 & 52.2 \tiny $\pm$ 0.3 &  & 53.3 \tiny $\pm$ 0.3 \\ % 220207A
1024 & & 83.6 \tiny $\pm$ 0.2 & 87.3 \tiny $\pm$ 0.3 &  & 43.2 \tiny $\pm$ 0.6 & 56.3 \tiny $\pm$ 0.5 &  & 56.5 \tiny $\pm$ 0.2 \\ % 220207A
2048 & & 86.5 \tiny $\pm$ 0.2 & 89.8 \tiny $\pm$ 0.1 &  & 48.2 \tiny $\pm$ 1.1 & 60.5 \tiny $\pm$ 0.4 &  & 59.4 \tiny $\pm$ 0.3 \\ % 220207A
4096 & & 88.4 \tiny $\pm$ 0.1 & 91.5 \tiny $\pm$ 0.1 &  & 53.1 \tiny $\pm$ 0.5 & 64.0 \tiny $\pm$ 0.3 &  & 62.0 \tiny $\pm$ 0.4 \\ % 220207A
    \bottomrule
    \end{tabular}
    \caption{Detailed results of Figure~\ref{figure1_max_sequence_length_vs_f1}: the effectiveness of Longformer on the MIMIC-III development set.}
    \label{table-detailed-figure-1}
\end{table*}

\begin{table*}[t]
    \centering
    %\small
    %\setlength{\tabcolsep}{2pt} % Default value: 6pt
    \begin{tabular}{rc cc c cc cc}
    \toprule
    & & \multicolumn{2}{c}{AUC} & & \multicolumn{2}{c}{$F_1$} & & \\
    \cmidrule{3-4} \cmidrule{6-7}
    & & Macro & Micro & & Macro & Micro & & P@5 \\
    \midrule
    \multicolumn{9}{c}{Longformer on \textbf{MIMIC-III}} \\
    \midrule
Vanilla & & 88.4 \tiny $\pm$ 0.1 & 91.5 \tiny $\pm$ 0.1 & & 53.1 \tiny $\pm$ 0.5 & 64.0 \tiny $\pm$ 0.3 & & 62.0 \tiny $\pm$ 0.4 \\ % 2022-02-07-A
TAPT & & 90.3 \tiny $\pm$ 0.2 & 92.7 \tiny $\pm$ 0.1 & & 60.8 \tiny $\pm$ 0.4 & 68.5 \tiny $\pm$ 0.3 & & 64.8 \tiny $\pm$ 0.3 \\ % 2022-02-07-A
    \midrule
    \multicolumn{9}{c}{RoBERTa on \textbf{MIMIC-III}} \\
    \midrule
Vanilla & & 81.6 \tiny $\pm$ 0.2 & 85.0 \tiny $\pm$ 0.3 & & 43.2 \tiny $\pm$ 1.7 & 53.9 \tiny $\pm$ 0.4 & & 54.0 \tiny $\pm$ 0.2 \\ % 2022-03-02-D
TAPT & & 82.3 \tiny $\pm$ 0.4 & 85.5 \tiny $\pm$ 0.3 & & 48.8 \tiny $\pm$ 0.4 & 56.7 \tiny $\pm$ 0.2 & & 55.3 \tiny $\pm$ 0.2 \\ % 2022-03-02-D
    \midrule
    \multicolumn{9}{c}{Longformer on \textbf{ECtHR}} \\
    \midrule
Vanilla & & --- & --- & & 77.4 \tiny $\pm$ 2.3 & 81.3 \tiny $\pm$ 0.3 & & --- \\ % 2022-02-07-A
TAPT & & --- & --- & & 78.5 \tiny $\pm$ 2.2 & 82.1 \tiny $\pm$ 0.6 & & --- \\ % 2022-02-07-A
    \midrule
    \multicolumn{9}{c}{RoBERTa on \textbf{ECtHR}} \\
    \midrule
Vanilla & & --- & --- & & 72.2 \tiny $\pm$ 1.5 & 74.8 \tiny $\pm$ 0.4 & & --- \\ % 2022-03-02-D
TAPT & & --- & --- & & 72.7 \tiny $\pm$ 0.7 & 75.1 \tiny $\pm$ 0.4 & & --- \\ % 2022-03-02-D
    \bottomrule
    \end{tabular}
    \caption{Detailed results of Figure~\ref{figure-tapt-results}: the impact of task-adaptive pre-training. Note that we use maximum sequence length $512$ for RoBERTa and $4096$ for Longformer in these experiments.}
    \label{table-detailed-figure-2}
\end{table*}

\begin{table*}[ht]
    \centering
    %\small
    %\setlength{\tabcolsep}{2pt} % Default value: 6pt
    \begin{tabular}{rc cc c cc cc cc}
    \toprule
    & & \multicolumn{2}{c}{AUC} & & \multicolumn{2}{c}{$F_1$} & & &\\
    \cmidrule{3-4} \cmidrule{6-7}
    Size & & Macro & Micro & & Macro & Micro & & P@5 & & Accuracy \\
    \midrule
    \multicolumn{11}{c}{\textbf{MIMIC-III}} \\
    \midrule
32 && 89.8 \tiny $\pm$ 0.1 & 92.3 \tiny $\pm$ 0.1 & & 59.6 \tiny $\pm$ 0.6 & 67.9 \tiny $\pm$ 0.3 && 64.2 \tiny $\pm$ 0.3 && --- \\ % 2022-02-07-A
64 && 90.0 \tiny $\pm$ 0.1 & 92.5 \tiny $\pm$ 0.1 & & 60.3 \tiny $\pm$ 0.3 & 68.1 \tiny $\pm$ 0.1 && 64.5 \tiny $\pm$ 0.1 && --- \\ % 2022-02-07-A
128 && 90.1 \tiny $\pm$ 0.1 & 92.6 \tiny $\pm$ 0.1 & & 60.5 \tiny $\pm$ 0.7 & 68.3 \tiny $\pm$ 0.3 && 64.7 \tiny $\pm$ 0.3 && --- \\ % 2022-02-07-A
256 && 90.2 \tiny $\pm$ 0.0 & 92.6 \tiny $\pm$ 0.1 & & 60.7 \tiny $\pm$ 0.6 & 68.4 \tiny $\pm$ 0.3 && 64.6 \tiny $\pm$ 0.2 && --- \\ % 2022-02-07-A
512 && 90.3 \tiny $\pm$ 0.2 & 92.7 \tiny $\pm$ 0.1 & & 60.8 \tiny $\pm$ 0.4 & 68.5 \tiny $\pm$ 0.3 && 64.8 \tiny $\pm$ 0.3 && --- \\ % 2022-02-07-A
    \midrule
    \multicolumn{11}{c}{\textbf{ECtHR}} \\
    \midrule
32 && --- & --- & & 78.2 \tiny $\pm$ 1.2 & 81.2 \tiny $\pm$ 0.3 && --- && --- \\ % 2022-02-07-A
64 && --- & --- & & 78.6 \tiny $\pm$ 1.7 & 81.4 \tiny $\pm$ 0.1 && --- && --- \\ % 2022-02-07-A
128 && --- & --- & & 79.9 \tiny $\pm$ 1.6 & 82.1 \tiny $\pm$ 0.5 && --- && --- \\ % 2022-02-07-A
256 && --- & --- & & 78.5 \tiny $\pm$ 2.1 & 81.8 \tiny $\pm$ 0.4 && --- && --- \\ % 2022-02-07-A
512 && --- & --- & & 78.5 \tiny $\pm$ 2.2 & 82.1 \tiny $\pm$ 0.6 && --- && --- \\ % 2022-02-07-A
    \midrule
    \multicolumn{11}{c}{\textbf{Hyperpartisan}} \\
    \midrule
32 && --- & --- & & -- & -- && --- && 83.9 \tiny $\pm$ 0.7 \\ % 2022-02-07-A
64 && --- & --- & & -- & -- && --- && 83.3 \tiny $\pm$ 1.9 \\ % 2022-02-07-A
128 && --- & --- & & -- & -- && --- && 83.9 \tiny $\pm$ 0.7 \\ % 2022-02-07-A
256 && --- & --- & & -- & -- && --- && 88.0 \tiny $\pm$ 0.7 \\ % 2022-02-07-A
512 && --- & --- & & -- & -- && --- && 85.9 \tiny $\pm$ 2.2 \\ % 2022-02-07-A
    \midrule
    \multicolumn{11}{c}{\textbf{20 News}} \\
    \midrule
32 && --- & --- & & -- & -- && --- && 92.8 \tiny $\pm$ 0.6 \\ % 2022-02-07-A
64 && --- & --- & & -- & -- && --- && 94.0 \tiny $\pm$ 0.5 \\ % 2022-02-07-A
128 && --- & --- & & -- & -- && --- && 93.8 \tiny $\pm$ 0.3 \\ % 2022-02-07-A
256 && --- & --- & & -- & -- && --- && 93.5 \tiny $\pm$ 0.1 \\ % 2022-02-07-A
512 && --- & --- & & -- & -- && --- && 94.0 \tiny $\pm$ 0.1 \\ % 2022-02-07-A
    \bottomrule
    \end{tabular}
    \caption{The impact of local attention window size in Longformer, measured on the development sets.}
    \label{appendix-table-window-in-Longformer}
\end{table*}

\begin{table*}[ht]
    \centering
    %\small
    %\setlength{\tabcolsep}{2pt} % Default value: 6pt
    \begin{tabular}{r cc c cc c}
    \toprule
    & \multicolumn{2}{c}{AUC} & & \multicolumn{2}{c}{$F_1$} & \\
    \cmidrule{2-3} \cmidrule{5-6}
    \# tokens & Macro & Micro & & Macro & Micro & P@5 \\
    \midrule
    \multicolumn{7}{c}{\textbf{MIMIC-III}} \\
    \midrule
1 & 90.1 \tiny $\pm$ 0.2 & 92.6 \tiny $\pm$ 0.1 & & 60.5 \tiny $\pm$ 0.9 & 68.2 \tiny $\pm$ 0.3 & 64.7 \tiny $\pm$ 0.3 \\ % 2021-10-24-C
8 & 90.0 \tiny $\pm$ 0.1 & 92.5 \tiny $\pm$ 0.1 & & 60.5 \tiny $\pm$ 0.7 & 68.2 \tiny $\pm$ 0.3 & 64.6 \tiny $\pm$ 0.2 \\ % 2021-10-24-C
16 & 90.0 \tiny $\pm$ 0.2 & 92.5 \tiny $\pm$ 0.1 & & 60.0 \tiny $\pm$ 0.2 & 68.1 \tiny $\pm$ 0.2 & 64.3 \tiny $\pm$ 0.3 \\ % 2021-10-24-C
32 & 90.0 \tiny $\pm$ 0.2 & 92.4 \tiny $\pm$ 0.1 & & 60.1 \tiny $\pm$ 0.5 & 67.9 \tiny $\pm$ 0.1 & 64.4 \tiny $\pm$ 0.2 \\ % 2021-10-24-C
64 & 89.9 \tiny $\pm$ 0.2 & 92.4 \tiny $\pm$ 0.1 & & 59.9 \tiny $\pm$ 1.0 & 67.9 \tiny $\pm$ 0.4 & 64.4 \tiny $\pm$ 0.3 \\ % 2021-10-24-C
    \midrule
    \multicolumn{7}{c}{\textbf{ECtHR}} \\
    \midrule
1 & --- & --- & & 78.5 \tiny $\pm$ 1.8 & 80.8 \tiny $\pm$ 0.4 & --- \\ % 2021-10-24-C
8 & --- & --- & & 77.2 \tiny $\pm$ 2.0 & 80.8 \tiny $\pm$ 0.4 & --- \\ % 2021-10-24-C
16 & --- & --- & & 77.7 \tiny $\pm$ 0.4 & 80.7 \tiny $\pm$ 0.3 & --- \\ % 2021-10-24-C
32 & --- & --- & & 78.2 \tiny $\pm$ 1.4 & 80.6 \tiny $\pm$ 0.4 & --- \\ % 2021-10-24-C
64 & --- & --- & & 77.7 \tiny $\pm$ 2.3 & 80.7 \tiny $\pm$ 0.5 & --- \\ % 2021-10-24-C
    \bottomrule
    \end{tabular}
    \caption{Detailed results of Figure~\ref{figure-global-attention-results}: the effect of applying global attention on more tokens, which are evenly chosen based on their positions.}
    \label{appendix-table-global-attention}
\end{table*}

\begin{table*}[ht]
    \centering
    %\small
    %\setlength{\tabcolsep}{2pt} % Default value: 6pt
    \begin{tabular}{r cc c cc c}
    \toprule
    & \multicolumn{2}{c}{AUC} & & \multicolumn{2}{c}{$F_1$} & \\
    \cmidrule{2-3} \cmidrule{5-6}
    \# tokens & Macro & Micro & & Macro & Micro & P@5 \\
    \midrule
    \multicolumn{7}{c}{\textbf{MIMIC-III}} \\
    \midrule
1 & 90.1 \tiny $\pm$ 0.2 & 92.6 \tiny $\pm$ 0.1 & & 60.5 \tiny $\pm$ 0.9 & 68.2 \tiny $\pm$ 0.3 & 64.7 \tiny $\pm$ 0.3 \\ % 2021-10-24-C
8 & 89.7 \tiny $\pm$ 0.2 & 92.0 \tiny $\pm$ 0.1 & & 61.0 \tiny $\pm$ 1.3 & 66.9 \tiny $\pm$ 0.4 & 64.0 \tiny $\pm$ 0.4 \\ % 2021-10-02-A
16 & 89.4 \tiny $\pm$ 0.2 & 91.9 \tiny $\pm$ 0.1 & & 60.1 \tiny $\pm$ 1.2 & 66.5 \tiny $\pm$ 0.3 & 63.9 \tiny $\pm$ 0.5 \\ % 2021-10-02-A
32 & 89.4 \tiny $\pm$ 0.4 & 91.9 \tiny $\pm$ 0.2 & & 60.3 \tiny $\pm$ 1.6 & 66.4 \tiny $\pm$ 0.6 & 63.7 \tiny $\pm$ 0.7 \\ % 2021-10-02-A
64 & 89.1 \tiny $\pm$ 0.4 & 91.7 \tiny $\pm$ 0.2 & & 59.4 \tiny $\pm$ 2.0 & 66.2 \tiny $\pm$ 0.7 & 63.4 \tiny $\pm$ 0.7 \\ % 2021-10-02-A
    \midrule
    \multicolumn{7}{c}{\textbf{ECtHR}} \\
    \midrule
1 & --- & --- & & 78.5 \tiny $\pm$ 1.8 & 80.8 \tiny $\pm$ 0.4 & --- \\ % 2021-10-24-C
8 & --- & --- & & 79.2 \tiny $\pm$ 0.3 & 80.9 \tiny $\pm$ 0.2 & --- \\ % 2021-11-10-A
16 & --- & --- & & 77.6 \tiny $\pm$ 1.2 & 80.4 \tiny $\pm$ 0.4 & --- \\ % 2021-11-10-A
32 & --- & --- & & 77.1 \tiny $\pm$ 0.7 & 80.0 \tiny $\pm$ 0.2 & --- \\ % 2021-11-10-A
64 & --- & --- & & 76.6 \tiny $\pm$ 1.1 & 79.9 \tiny $\pm$ 0.5 & --- \\ % 2021-11-10-A
    \bottomrule
    \end{tabular}
    \caption{The effect of applying global attention on more informative tokens, which are identified based on TF-IDF.}
    \label{appendix-table-global-attention-content-based}
\end{table*}

\begin{table*}[ht]
    \centering
    \small
    \begin{tabular}{r c cc c cc cc cc}
    \toprule
    && \multicolumn{2}{c}{AUC} & & \multicolumn{2}{c}{$F_1$} && \\
    \cmidrule{3-4} \cmidrule{6-7}
    Size && Macro & Micro & & Macro & Micro && P@5 && Accuracy \\
    \midrule
    \multicolumn{11}{c}{Disjoint segments on \textbf{MIMIC-III}} \\
    \midrule
%32 & 89.4 \tiny $\pm$ 0.1 & 92.1 \tiny $\pm$ 0.0 & & 60.8 \tiny $\pm$ 0.5 & 67.7 \tiny $\pm$ 0.2 & 63.3 \tiny $\pm$ 0.2 \\ % 2021-09-20-A
64 && 89.4 \tiny $\pm$ 0.1 & 92.0 \tiny $\pm$ 0.1 & & 60.8 \tiny $\pm$ 1.1 & 67.9 \tiny $\pm$ 0.3 && 63.5 \tiny $\pm$ 0.3 && --- \\ % 2021-09-20-A
128 && 89.5 \tiny $\pm$ 0.1 & 92.1 \tiny $\pm$ 0.1 & & 61.2 \tiny $\pm$ 0.6 & 68.0 \tiny $\pm$ 0.3 && 63.5 \tiny $\pm$ 0.3 && --- \\ % 2021-09-20-A
256 && 89.6 \tiny $\pm$ 0.1 & 92.1 \tiny $\pm$ 0.1 & & 61.0 \tiny $\pm$ 0.4 & 67.6 \tiny $\pm$ 0.2 && 63.6 \tiny $\pm$ 0.2 && --- \\ % 2021-09-20-A
512 && 89.2 \tiny $\pm$ 0.2 & 91.8 \tiny $\pm$ 0.2 & & 59.4 \tiny $\pm$ 0.5 & 66.7 \tiny $\pm$ 0.3 && 63.4 \tiny $\pm$ 0.4 && --- \\ % 2021-09-20-A
    \midrule
    \multicolumn{11}{c}{Overlapping segments on \textbf{MIMIC-III}} \\
    \midrule
%32 & 89.7 \tiny $\pm$ 0.2 & 92.3 \tiny $\pm$ 0.1 & & 61.7 \tiny $\pm$ 0.3 & 68.2 \tiny $\pm$ 0.4 & 63.7 \tiny $\pm$ 0.1 \\ % 2021-10-11-A
64 && 89.7 \tiny $\pm$ 0.1 & 92.3 \tiny $\pm$ 0.1 & & 62.3 \tiny $\pm$ 0.2 & 68.7 \tiny $\pm$ 0.1 && 64.1 \tiny $\pm$ 0.1 && --- \\ % 2021-10-11-A
128 && 89.7 \tiny $\pm$ 0.2 & 92.3 \tiny $\pm$ 0.1 & & 61.8 \tiny $\pm$ 0.9 & 68.5 \tiny $\pm$ 0.3 && 64.0 \tiny $\pm$ 0.2 && --- \\ % 2021-10-11-A
256 && 89.5 \tiny $\pm$ 0.1 & 92.1 \tiny $\pm$ 0.1 & & 61.4 \tiny $\pm$ 0.3 & 68.1 \tiny $\pm$ 0.2 && 63.8 \tiny $\pm$ 0.1 && --- \\ % 2021-10-11-A
512 && 89.4 \tiny $\pm$ 0.1 & 92.0 \tiny $\pm$ 0.0 & & 60.3 \tiny $\pm$ 0.3 & 67.2 \tiny $\pm$ 0.2 && 63.6 \tiny $\pm$ 0.3 && --- \\ % 2021-10-11-A
    \midrule
    \multicolumn{11}{c}{Disjoint segments on \textbf{ECtHR}} \\
    \midrule
%32 & --- & --- & & 75.5 \tiny $\pm$ 1.7 & 79.3 \tiny $\pm$ 0.4 & --- \\ % 2021-10-13-C
64 && --- & --- & & 76.6 \tiny $\pm$ 1.2 & 79.7 \tiny $\pm$ 0.2 && --- && --- \\ % 2021-10-13-C
128 && --- & --- & & 77.6 \tiny $\pm$ 2.3 & 80.8 \tiny $\pm$ 0.4 && --- && --- \\ % 2021-10-13-C
256 && --- & --- & & 77.7 \tiny $\pm$ 1.4 & 81.2 \tiny $\pm$ 0.4 && --- && --- \\ % 2021-10-13-C
512 && --- & --- & & 78.3 \tiny $\pm$ 1.3 & 81.7 \tiny $\pm$ 0.3 && --- && --- \\ % 2021-10-13-C
    \midrule
    \multicolumn{11}{c}{Overlapping segments on \textbf{ECtHR}} \\
    \midrule
%32 & --- & --- & & 74.1 \tiny $\pm$ 2.6 & 79.4 \tiny $\pm$ 0.6 & --- \\ % 2021-10-18-C
64 && --- & --- & & 76.9 \tiny $\pm$ 1.7 & 80.5 \tiny $\pm$ 0.5 && --- && --- \\ % 2021-10-18-C
128 && --- & --- & & 77.5 \tiny $\pm$ 1.7 & 81.2 \tiny $\pm$ 0.5 && --- && --- \\ % 2021-10-18-C
256 && --- & --- & & 78.1 \tiny $\pm$ 1.4 & 81.5 \tiny $\pm$ 0.2 && --- && --- \\ % 2021-10-18-C
512 && --- & --- & & 78.4 \tiny $\pm$ 1.5 & 81.4 \tiny $\pm$ 0.4 && ---  && ---\\ % 2021-10-18-C
    \midrule
    \multicolumn{11}{c}{Disjoint segments on \textbf{Hyperpartisan}} \\
    \midrule
64 && --- & --- & & --- & --- && --- && 88.8 \tiny $\pm$ 1.8 \\ % 2022-03-12-B
128 && --- & --- & & --- & --- && --- && 89.1 \tiny $\pm$ 1.4 \\ % 2022-03-12-B
256 && --- & --- & & --- & --- && --- && 87.8 \tiny $\pm$ 1.8 \\ % 2022-03-12-B
512 && --- & --- & & --- & --- && --- && 86.2 \tiny $\pm$ 1.8 \\ % 2022-03-12-B
    \midrule
    \multicolumn{11}{c}{Overlapping segments on \textbf{Hyperpartisan}} \\
    \midrule
64 && --- & --- & & --- & --- && --- && 87.5 \tiny $\pm$ 1.4 \\ % 2022-03-12-B
128 && --- & --- & & --- & --- && --- && 88.4 \tiny $\pm$ 1.2 \\ % 2022-03-12-B
256 && --- & --- & & --- & --- && --- && 88.1 \tiny $\pm$ 2.1 \\ % 2022-03-12-B
512 && --- & --- & & --- & --- && --- && 88.4 \tiny $\pm$ 0.8 \\ % 2022-03-12-B
    \midrule
    \multicolumn{11}{c}{Disjoint segments on \textbf{20 News}} \\
    \midrule
64 && --- & --- & & --- & --- && --- && 93.3 \tiny $\pm$ 0.2 \\ % 2022-03-12-B
128 && --- & --- & & --- & --- && --- && 93.5 \tiny $\pm$ 0.3 \\ % 2022-03-12-B
256 && --- & --- & & --- & --- && --- && 94.4 \tiny $\pm$ 0.4 \\ % 2022-03-12-B
512 && --- & --- & & --- & --- && --- && 94.0 \tiny $\pm$ 0.3 \\ % 2022-03-12-B
    \midrule
    \multicolumn{11}{c}{Overlapping segments on \textbf{20 News}} \\
    \midrule
64 && --- & --- & & --- & --- && --- && 93.8 \tiny $\pm$ 0.4 \\ % 2022-03-12-B
128 && --- & --- & & --- & --- && --- && 93.4 \tiny $\pm$ 0.3 \\ % 2022-03-12-B
256 && --- & --- & & --- & --- && --- && 94.5 \tiny $\pm$ 0.2 \\ % 2022-03-12-B
512 && --- & --- & & --- & --- && --- && 93.9 \tiny $\pm$ 0.3 \\ % 2022-03-12-B
    \bottomrule
    \end{tabular}
    \caption{The effect of varying the segment length and whether allowing segments to overlap in the hierarchical transformers.}
    \label{appendix-table-segment-length-no-overlap}
\end{table*}

\begin{table*}[ht]
    \centering
    %\small
    %\setlength{\tabcolsep}{2pt} % Default value: 6pt
    \begin{tabular}{r cc c cc c}
    \toprule
    & \multicolumn{2}{c}{AUC} & & \multicolumn{2}{c}{$F_1$} & \\
    \cmidrule{2-3} \cmidrule{5-6}
    & Macro & Micro & & Macro & Micro & P@5 \\
    \midrule
    \multicolumn{7}{c}{\textbf{MIMIC-III}} \\
    \midrule
E (4096) & 89.7 \tiny $\pm$ 0.2 & 92.3 \tiny $\pm$ 0.1 & & 61.8 \tiny $\pm$ 0.9 & 68.5 \tiny $\pm$ 0.3 & 64.0 \tiny $\pm$ 0.2 \\ % 2021-10-11-A
S (4096) & 87.2 \tiny $\pm$ 0.2 & 90.1 \tiny $\pm$ 0.2 & & 55.2 \tiny $\pm$ 0.4 & 62.9 \tiny $\pm$ 0.2 & 59.9 \tiny $\pm$ 0.2 \\ % 2021-11-07-A
S (6144) & 88.2 \tiny $\pm$ 0.2 & 91.0 \tiny $\pm$ 0.2 & & 57.8 \tiny $\pm$ 0.3 & 65.4 \tiny $\pm$ 0.3 & 61.7 \tiny $\pm$ 0.3 \\ % 2021-11-07-A
S (8192) & 88.5 \tiny $\pm$ 0.3 & 91.2 \tiny $\pm$ 0.2 & & 58.8 \tiny $\pm$ 0.2 & 66.0 \tiny $\pm$ 0.4 & 62.4 \tiny $\pm$ 0.1 \\ % 2021-11-07-A
    \midrule
    \multicolumn{7}{c}{\textbf{ECtHR}} \\
    \midrule
E (4096) & --- & --- & & 77.5 \tiny $\pm$ 1.7 & 81.2 \tiny $\pm$ 0.5 & --- \\ % 2021-10-18-C
S (4096) & --- & --- & & 75.3 \tiny $\pm$ 1.3 & 80.1 \tiny $\pm$ 0.4 & --- \\ % 2021-11-07-A
S (6144) & --- & --- & & 77.1 \tiny $\pm$ 1.8 & 80.5 \tiny $\pm$ 0.5 & --- \\ % 2021-11-07-A
S (8192) & --- & --- & & 77.7 \tiny $\pm$ 1.9 & 81.3 \tiny $\pm$ 0.5 & --- \\ % 2021-11-07-A
    \bottomrule
    \end{tabular}
    \caption{Detailed results of Figure~\ref{figure-sentence_boundary_vs_evenly_splitting}: a comparison between evenly splitting and splitting based on document structure. E: evenly splitting; S: splitting based on document structure.}
    \label{appendix-table-boundary-splitting}
\end{table*}

\begin{table*}[ht]
    \centering
    %\small
    %\setlength{\tabcolsep}{2pt} % Default value: 6pt
    \begin{tabular}{r cc c cc c}
    \toprule
    & \multicolumn{2}{c}{AUC} & & \multicolumn{2}{c}{$F_1$} & \\
    \cmidrule{2-3} \cmidrule{5-6}
    & Macro & Micro & & Macro & Micro & P@5 \\
    \midrule
    \multicolumn{7}{c}{\textbf{MIMIC-III}} \\
    \midrule
Longformer & 90.0 \tiny $\pm$ 0.2 & 92.5 \tiny $\pm$ 0.1 & & 60.0 \tiny $\pm$ 0.2 & 68.1 \tiny $\pm$ 0.2 & 64.3 \tiny $\pm$ 0.3 \\ % 2021-10-24-C
 + LWAN & 90.5 \tiny $\pm$ 0.2 & 92.9 \tiny $\pm$ 0.2 & & 62.2 \tiny $\pm$ 0.7 & 69.2 \tiny $\pm$ 0.3 & 65.1 \tiny $\pm$ 0.1 \\ % 2021-11-11-A
Hierarchical & 89.7 \tiny $\pm$ 0.2 & 92.3 \tiny $\pm$ 0.1 & & 61.8 \tiny $\pm$ 0.9 & 68.5 \tiny $\pm$ 0.3 & 64.0 \tiny $\pm$ 0.2 \\ % 2021-10-11-A
 + LWAN & 91.4 \tiny $\pm$ 0.1 & 93.7 \tiny $\pm$ 0.1 & & 64.2 \tiny $\pm$ 0.4 & 70.3 \tiny $\pm$ 0.1 & 65.3 \tiny $\pm$ 0.1 \\ % 2021-11-06-A
    \midrule
    \multicolumn{7}{c}{\textbf{ECtHR}} \\
    \midrule
Longformer & --- & --- & & 77.7 \tiny $\pm$ 0.4 & 80.7 \tiny $\pm$ 0.3 & --- \\ % 2021-11-03-A
 + LWAN & --- & --- & & 79.5 \tiny $\pm$ 0.8 & 81.1 \tiny $\pm$ 0.3 & --- \\ % 2021-11-11-A
Hierarchical & --- & --- & & 77.5 \tiny $\pm$ 1.7 & 81.2 \tiny $\pm$ 0.5 & --- \\ % 2021-10-18-C
 + LWAN & --- & --- & & 79.7 \tiny $\pm$ 0.9 & 81.3 \tiny $\pm$ 0.3 & --- \\ % 2021-11-06-A
    \bottomrule
    \end{tabular}
    \caption{The effect of label-wise attention network.}
    \label{appendix-table-hierarchical-lwan}
\end{table*}